\documentclass{ecai} 

\usepackage{latexsym}
\usepackage{amssymb}
\usepackage{amsmath}
\usepackage{amsthm}
\usepackage{booktabs}
\usepackage{enumitem}
\usepackage{graphicx}
\usepackage{color}
\usepackage{lipsum}
\usepackage{algorithm}
\usepackage{algpseudocode}
\usepackage{microtype}
\usepackage{float}
\usepackage{booktabs}
\usepackage{placeins}
\usepackage{subfigure}
\usepackage{bm}
\usepackage{changepage}
\usepackage{fancyhdr}
\usepackage{url}

\newcommand{\BibTeX}{B\kern-.05em{\sc i\kern-.025em b}\kern-.08em\TeX}

\begin{document}


\begin{frontmatter}


\paperid{7400} 


\title{Pareto-NRPA: A Novel Monte-Carlo Search Algorithm for Multi-Objective Optimization}


\author[A, B]{\fnms{Noé}~\snm{Lallouet} \thanks{Corresponding Author. Email: noe.lallouet@dauphine.eu.}}
\author[A]{\fnms{Tristan}~\snm{Cazenave}}
\author[B]{\fnms{Cyrille}~\snm{Enderli}} 

\address[A]{LAMSADE, Paris Dauphine - PSL University, Paris, France}
\address[B]{Thales DMS, Elancourt, France}


\begin{abstract}
We introduce Pareto-NRPA, a new Monte-Carlo algorithm designed for multi-objective optimization problems over discrete search spaces. Extending the Nested Rollout Policy Adaptation (NRPA) algorithm originally formulated for single-objective problems, Pareto-NRPA generalizes the nested search and policy update mechanism to multi-objective optimization. The algorithm uses a set of policies to concurrently explore different regions of the solution space and maintains non-dominated fronts at each level of search. Policy adaptation is performed with respect to the diversity and isolation of sequences within the Pareto front. We benchmark Pareto-NRPA on two classes of problems: a novel bi-objective variant of the Traveling Salesman Problem with Time Windows problem (MO-TSPTW), and a neural architecture search task on well-known benchmarks. Results demonstrate that Pareto-NRPA achieves competitive performance against state-of-the-art multi-objective algorithms, both in terms of convergence and diversity of solutions. Particularly, Pareto-NRPA strongly outperforms state-of-the-art evolutionary multi-objective algorithms on constrained search spaces. To our knowledge, this work constitutes the first adaptation of NRPA to the multi-objective setting.
\end{abstract}

\end{frontmatter}


\section{Introduction}

Multi-objective optimization (MOO) is a research area consisting in finding ways to optimize several, often conflicting, objective functions. A MOO problem can be formulated in the following way: 
\begin{equation} \label{eq:moo}
    \begin{aligned}
    & \min_x (f_1(x), f_2(x), ..., f_p(x)) \\ 
    & \text{subject to } x \in X
    \end{aligned}
\end{equation}
where $(f_1, ..., f_p)$ are the objective functions and $X$ is the feasible set. $X$ can be a subset of $\mathbb{R}^n$ for continuous optimization problems or $\mathbb{N}^n$ for discrete optimization tasks, and includes possible constraints.  The concept of Pareto-dominance allows one to compare two solutions $x$ and $y$. It is said that a solution $x$ \textit{dominates} $y$ ($x \prec y$) iff $f_i(x) \leq f_i(y), \forall i=1, ..., k$ and there exists a $j \in {1, ..., p}$ such that $f_j(x) < f_j(y)$. A solution $x$ is said Pareto-optimal iff $\neg \exists y \in X: y \prec x$, i.e. if there is no solution $y$ that dominates $x$. The set of all Pareto-optimal solutions is called the Pareto front.

In this paper, we are concerned with tackling multi-objective optimization problems over a discrete search space using Monte-Carlo methods. Monte-Carlo methods typically use random simulations to collect information about the search space and converge to a solution.
We introduce Pareto-NRPA, which is an extension of the NRPA algorithm \cite{rosin_nested_2011} to the case of multi-objective optimization. NRPA is a Monte-Carlo algorithm tailored to single-objective optimization (SOO) that uses nested searches combined with policy updates. NRPA has shown high performance on games and combinatorial optimization problems, finding world records on Morpion Solitaire and crossword puzzles \cite{rosin_nested_2011}.
The contributions of this work are the following: 
\begin{itemize}
    \item We propose Pareto-NRPA, a novel algorithm for multi-objective optimization, and discuss the motivations behind the design choices of the algorithm. The code for Pareto-NRPA is available online. \footnote{\url{https://github.com/pareto-nrpa/pareto-nrpa}}
    \item We introduce a new benchmark dataset based on popular traveling Salesman Problem with Time Windows (TSPTW) instances adapted to MOO. This dataset (MO-TSPTW) is publicly released and available online. \footnote{\url{https://github.com/pareto-nrpa/mo-tsptw}}
    \item We compare the results of Pareto-NRPA against state-of-the-art MOO algorithms on two different sets of problems: the bi-objective TSPTW problem defined previously and bi-objective Neural Architecture Search (NAS) on well-known NAS benchmark datasets.
\end{itemize}
We find that Pareto-NRPA exhibits strong performances on these problems. As such, Pareto-NRPA may be identified as a suitable candidate for many multi-objective optimization problems over discrete search spaces. Similarly to evolutionary multi-objective optimization algorithms, Pareto-NRPA is a non-exact method for multi-objective optimization.
To the best of our knowledge, this work is the first generalization of the NRPA algorithm to the multi-objective optimization setting.


\section{Related Works}

\subsection*{Multi-objective optimization}

A large part of the literature in MOO is related to evolutionary computing, and particularly genetic algorithms.
Multi-objective evolutionary algorithms (MOEAs) typically generate good solutions while keeping a large degree of diversity in the solution set. 
NSGA-II \cite{deb_fast_2002} is one of such algorithms, and is popular in MOO due to its ease of use and adaptability. Other popular genetic algorithms for multi-objective optimization include SMS-EMOA \cite{hochstrate_sms-emoa_2007}, SPEA-2 \cite{zitzler_spea2_2001}, NSGA-III \cite{deb_evolutionary_2014} and MOEA/D \cite{zhang_moead_2007}. MOEAs have been used successfully for many multi-objective optimization problems, such as water management \cite{lewis_solving_2017} or change detection in satellite images \cite{yavariabdi_change_2017}.

Approaches to MOO using reinforcement learning have also been proposed \cite{van_moffaert_hypervolume-based_2013}.
Multi-objective reinforcement learning (MORL) is a branch of reinforcement learning that aims to learn a set of policies for multi-objective Markov decision processes. MORL has been successfully applied to many real-world problems \cite{oh_multi-objective_2024} \cite{li_urban_2019} \cite{castelletti_tree-based_2012}.
Search and planning methods such as Monte-Carlo Tree Search, on the other hand, are less popular approaches for MOO. The UCT algorithm \cite{kocsis_bandit_2006}, one of the foremost MCTS algorithms, has notably been adapted to MOO \cite{chen_pareto_2019}, \cite{wang_multi-objective_nodate}. While there are similarities between RL and MCTS techniques \cite{vodopivec_monte_2017}, MCTS algorithms largely ignore the discounted reward mechanism, which is crucial in RL, in favour of terminal rewards observed after a playout. MCTS methods also often keep in memory only part of the search space and collect information through simulated playouts instead of real interaction with the environment.

\subsection*{Monte-Carlo Tree Search and NRPA}

Monte-Carlo Tree Search (MCTS) is a popular set of techniques used to explore a search tree. MCTS uses random playouts to collect information about the value of a node in the search tree and uses a selection formula to balance exploration and exploitation in the tree.  The UCT algorithm \cite{kocsis_bandit_2006} and RAVE \cite{gelly_monte-carlo_2011} are some node selection methods. 

Nested Monte-Carlo Search (NMCS) \cite{cazenave_nested_2009} is a different way of exploring a search tree. NMCS implements a nested structure, which means that each search recursively launches a lower-level search, until level 0 where the search returns the terminal reward of a random playout starting from the current state. A state is a point in the search space representing a partial (in which case, actions can be taken) or terminal solution. Starting from an empty sequence, NMCS chooses actions until it reaches a terminal state. In state $s_i$, a level $\ell$ NMCS search launches a level $\ell-1$ search for each available action. The action with the highest score is played. NMCS has been successfully applied to various games and combinatorial optimization problems \cite{roucairol_solving_2023} \cite{cazenave_nested_2016}.

NRPA \cite{rosin_nested_2011} combines the nested structure of NMCS with a policy learning method. At each level, NRPA performs a lower-level search and updates the policy with respect to the best returned sequence. Each search of level $\ell$ launches $n$ searches of level $\ell - 1$. At level 0, a search is a playout where the actions are conditioned by a policy vector $\pi$. The policy $\pi$ associates a weight $w_{s, a}$ to a $(state, action)$ couple. $(state, action)$ couples are uniquely encoded by a domain-specific $code$ method. 

During a playout and at state $s_i$, the probability of picking action $k$ is derived from the associated policy weight: $p_{ik} = \frac{e^{w_{i,k}}}{\sum_j e^{w_{i,j}}}$. At algorithm initialization, $\pi$ is a random policy with uniform distribution over the actions. After each search, the policy $\pi$ is updated using gradient ascent towards the best sequence found at the current level. 
Algorithms \ref{alg:playout} shows the playout algorithm. The reader is kindly referred to the original NRPA paper \cite{rosin_nested_2011} for pseudocodes of the NRPA algorithm and the Adapt algorithm.

\begin{algorithm}[t]
    \caption{Playout} \label{alg:playout}
    \begin{algorithmic}[1]
        \Require $state, \pi$
        \State $sequence \gets []$
        \While{$true$}
            \If{$state$ is terminal}
                \State \Return $(\text{score}(state), sequence)$
            \EndIf
            \State $z \gets 0$
            \For{$m \in \text{possible moves for }state$}
                \State $z \gets z + \exp (\pi[\text{code}(m)])$
            \EndFor
            \State choose a $move$ with probability $\frac{\exp (\pi[\text{code}(move)])} {z}$
            \State $state \gets \text{play}(state, move)$
            \State $sequence \gets sequence \cup move$
        \EndWhile
    \end{algorithmic}
\end{algorithm}

The research work of \cite{cornu_local_2017} introduces A-NMCS, a variant of NMCS for MOO problems, using scalarization to decompose a multi-objective problem into several single-objective problems, while highlighting that no adaptation of NRPA has been proposed for multi-objective optimization. We aim to fill this gap in the literature by proposing Pareto-NRPA. 


\section{Contributions} \label{sec:contributions}

Given the space of all possible sequences $S$, NRPA tries to find the sequence with the highest score $s^* = \arg \max_{s \in S} \text{score}(s)$. As such, NRPA is suited to single-objective optimization (SOO) problems. In the context of multi-objective optimization, one aims to identify, rather than a single sequence $s^*$, a set $S^*$ of Pareto-efficient sequences:
\begin{equation}
    S^* = \{s \in S | \neg \exists u \in S: u \neq s, u \prec s \}
\end{equation}

In order to optimize several objectives simultaneously, a number of changes have to be made to the original NRPA algorithm. First of all, instead of optimizing a single policy $\pi$, we introduce a set of policies $\Pi = \{\pi_1, ..., \pi_n\}$. The cardinality of $\Pi$ is a user-defined hyperparameter. By using several policies instead of one, the algorithm is able to optimize different regions of the search space, with one policy focusing on its own region. 

At the end of a search, the original NRPA algorithm returns the best score and its associated sequence. The \texttt{score()} method takes as input a solution and returns the objective function values for that solution. In Pareto-NRPA, they are replaced by the Pareto set of non-dominated scores and the associated sequences. For each playout (i.e. a level $0$ search), Pareto-NRPA first samples a policy $\pi_k \in \Pi$ from a uniform distribution, then uses the policy to guide a search starting from the root where probabilities of taking actions are conditioned by policy weights. When the playout reaches a terminal state, the objective functions $f_1, ..., f_p$ are evaluated and returned. For each sequence, the policy that was followed during the playout is memorized. Storing the sequence-policy couples $(s, \pi_k)$ can be done in a tabular fashion or (as implemented in this work) by storing $\pi_k$ as an attribute of the evaluated sequence. Line 2 of Algorithm \ref{alg:pareto-nrpa} indicates that the policy is sampled from a uniform distribution. Using a uniform distribution promotes a similar number of function evaluations for all policies. More complex policy sampling methods, such as weighting policies according to their relative performance, remain an avenue for future work.

After a level $\ell - 1$ search, instead of memorizing the best score and the associated sequence, Pareto-NRPA memorizes a non-dominated set. The algorithm uses non-dominated sorting \cite{deb_fast_2002} to categorize the result of a search in several fronts $F_0, F_1, ..., F_p$ and memorizes the solutions belonging to the dominating set $F_0$. Algorithm \ref{alg:pareto-adapt} shows that policies are adapted with respect to the solutions they have produced if and only if these solutions belong to $S^*$ (line 9). As such, if a policy has not led to a non-dominated solution, it will not be updated. In order to update each policy at least once, a minimum of one solution per policy is kept in $S^*$. If a policy $\pi_k$ is not represented in $F_0$,the next fronts $F_1, ... F_p$ are iterated through until the first solution belonging to $\pi_k$ is found.  

The Adapt algorithm in Pareto-NRPA differs from the original NRPA formulation by incorporating multiple sequences into the policy update, rather than a single best one. In multi-objective optimization, a solution may outperform another with respect to one objective while being inferior with respect to another. Thus, the policy update cannot rely on a single dominant sequence but must consider a set of non-dominated sequences. For each sequence $s \in S^*$ generated under policy $\pi_k$, the Adapt algorithm performs a weighted gradient update, conditioned by $\alpha$, which is a user-defined hyperparameter representing the learning rate. The weight assigned to each sequence depends on its relative isolation within the non-dominated front, which is calculated using the crowding distance (CD) \cite{deb_fast_2002}. This distance approximates the perimeter of the hyper-rectangle defined by the nearest neighbors of a point in objective space, thus providing a measure of how sparsely populated the region around a sequence is. 

Sequences with a high crowding distance are considered more isolated and are more likely to contribute to high diversity in the solution space. To prevent instability during policy updates, particularly for sequences with extreme objective function values that yield a $+ \infty$ CD, the values are clipped to a maximum of 2. This ensures that the policy is updated preferentially towards isolated, diverse sequences without causing gradient explosion. An ablation study evaluating Pareto-NRPA with and without weighting sequences using CD is presented in the supplementary material of this paper (Section \ref{sec:supplementary}, Table \ref{table:crowding-nrpa}).

The NRPA and Adapt algorithms adapted to the multi-objective optimization setting are presented in Algorithms \ref{alg:pareto-nrpa} and \ref{alg:pareto-adapt}.

\begin{algorithm}[h]
    \caption{Pareto-NRPA} \label{alg:pareto-nrpa}
    \begin{algorithmic}[1]
    \Require $level$, policy set $\Pi$
        \If{$level = 0$}
            \State Choose $\pi_k: p(\pi_k) \sim U[0, |\Pi|]$
            \State \Return Playout($root$, $\pi_k$)
        \Else
            \State $S^* \gets \emptyset$
            \For{$N$ iterations}
                \State $set \gets $Pareto-NRPA($level-1, \Pi$)
                \State $S^* \gets S^* \cup set$
                \State $S^* \gets \left\{ s \in S^* | \neg \exists u \in S^*: u \neq s, u \prec s\right\}$
                \State $\Pi \gets $Pareto-Adapt($\Pi, S^*$)
            \EndFor
            \State \Return $S^*$
        \EndIf
    \end{algorithmic}
\end{algorithm}

\begin{algorithm}[t]
    \caption{Pareto-Adapt} \label{alg:pareto-adapt}
    \begin{algorithmic}[1]
        \Require Policy vector $\Pi$, optimal set $S^*$
        \State $D \gets \text{CrowdingDistance}(S^*)$
        \For{$s \in S^*$}
            \State $\pi \gets \text{policy used to sample } s$
            \State $\pi' \gets \pi$
            \State $state \gets root$
            \For{$move \in s$}
                \State $\pi'[\text{code}(move)] \gets \pi'[\text{code}(move)] + (\alpha * D[s])$
                \State $z \gets 0$
                \For{$m \in \text{possible moves for } state$}
                    \State $z \gets z + \exp(\pi[\text{code}(m)]])$
                \EndFor
                \For{$m \in \text{possible moves for } state$}
                    \State $\pi'[\text{code}(m)] \gets \pi'[\text{code}(m)] -  $\\
                    \hspace{3.5cm} $(\alpha * \frac{\exp(\pi[\text{code}(m)])}{z}*D[s])$
                \EndFor
            \State $state \gets \text{play}(state, move)$
            \EndFor
            \State $\pi \gets \pi'$
        \EndFor
        \State Return $\Pi$
    \end{algorithmic}
\end{algorithm}

It is straightforward to note that Pareto-NRPA with $n\_objectives=1$ and $|\Pi|=1$ is equivalent to NRPA. As such, Pareto-NRPA is a generalization of NRPA. To the best of our knowledge, this work is the first generalization of NRPA to multi-objective optimization.

A large number of improvements have been proposed to NRPA over time \cite{dang_warm-starting_2023} \cite{sentuc_learning_2023} \cite{cazenave_generalized_2020}. Crucially, GNRPA \cite{cazenave_generalized_2020} generalizes the NRPA algorithm to include a domain-specific bias term. GNRPA has been shown to improve performance on several problems, including the TSPTW. GNRPA is very easily implemented in Pareto-NRPA, as it requires minimal modifications to the algorithm. Specifically, line 10 of Algorithm \ref{alg:pareto-adapt} is replaced by $z \gets z + \exp \left( \pi(state, m) + \beta(m) \right)$ and line 13 is replaced by $\pi'[\text{code}(m)] \gets \pi'[\text{code}(m)] -  (\alpha * \frac{\exp(\pi[\text{code}(m)]+ \beta(m))}{z}*D[s])$.  

It may also be noted that Pareto-NRPA is parallelizable. Although more complex ways to perform parallelization will be the object of future research, a straightforward way to parallelize Pareto-NRPA is by implementing playout parallelization, where a master process runs the main Pareto-NRPA algorithm and launches several playouts simultaneously with follower processes. This way of parallelizing Pareto-NRPA is reminiscent of Stabilized-NRPA \cite{cazenave_stabilized_2021}.

\section{Results} \label{sec:results}

We benchmark the performances of Pareto-NRPA on two problems. The first problem is an extension of classical instances of the traveling Salesman Problem with Time Windows (TSPTW) to bi-objective optimization. The second problem relates to finding neural network architectures in a large search space (NAS).

Comparing sets of results in a MOO setting is not as straightforward as in the single-objective case. A solution set $S_1$ may contain elements that outperform those in another set $S_2$ with respect to one objective $f_1$, while simultaneously being dominated or inferior with respect to other objectives $f_2,..., f_p$. We compare sets of solutions obtained by different algorithms in the following ways:
\begin{itemize}
    \item \textbf{Qualitative: } For bi-objective optimization problems, the best solutions found by each run of the algorithm are aggregated and their global dominating front is plotted. This graphical and qualitative comparison allows one to identify performing regions, Pareto front spread, and relative dominance. 
    \item \textbf{Quantitative: } We use the hypervolume metric \cite{guerreiro_hypervolume_2022} to compare the size of the region that the Pareto front dominates over a reference point.
    Let $Y^*$ be the objective space values of the elements in the Pareto set approximation $S^*$.
    The hypervolume metric is defined as volume of the space dominated by $Y^*$ and bounded by a reference point $r$. Formally, the hypervolume indicator is written:
    \begin{equation} \label{eq:hypervolume}
        HV(Y^*, r) = \lambda_m \left( \bigcup_{y \in Y^*} [y, r]\right)
    \end{equation}
    where $\lambda_m$ is the Lebesgue measure in $m$ dimensions.

    In order to evaluate solution diversity and Pareto front homogeneity, the overall spread (OS) and spacing metrics are used. The overall spread metric, introduced by \cite{wu_metrics_2001}, gives information on the extent of the Pareto front. It is calculated in the following way: 
    \begin{equation}
        OS(Y^*) = \prod_{i=1}^p \frac{\left| \max\limits_{y \in Y^*}y_i - \min\limits_{y \in Y^*}y_i \right|}{\left| \tilde{y}^l_i - \tilde{y}^M_i \right|}
    \end{equation} 
    where $\tilde{y}^l_i$ is an approximation of the ideal objective vector and $\tilde{y}^M_i$ is an approximation of the maximal objective vector.
    
    Finally, the spacing metric \cite{schott_fault_1995} returns a quantity related to the variation of the Manhattan distance between the elements of $Y^*$. Its expression is: 
    \begin{equation}
        SP(Y^*) = \sqrt{\frac{1}{|Y^* - 1|} \sum \limits_{j=1}^{|Y^*|} \left( \bar{d} - d^1(y^j, Y^* \setminus \{y^j\})^2  \right)}
    \end{equation}
    where $d^1(y^j, Y^* \setminus \{y^j\})$ is the minimal $L_1$-distance between an element $y^j$ and the rest of the elements of $Y^*$ and $\bar{d}$ is the mean of all $d^1(y^j, Y^* \setminus \{y^j\})$. A lower spacing value indicates a better distribution of points over the Pareto front approximation.
\end{itemize}

The results obtained by Pareto-NRPA are compared to the following algorithms:

\begin{itemize}
    \item NSGA-II \cite{deb_fast_2002} is one of the most popular multi-objective evolutionary algorithms. It uses non-dominated sorting and crowding distance assignment to manage a population of individuals. We run NSGA-II with a population size of 250 and a sample size of 25. 
    \item SMS-EMOA \cite{hochstrate_sms-emoa_2007} uses the hypervolume metric to as the selection criterion for survival. It is a state-of-the-art multi-objective evolutionary algorithm. SMS-EMOA is run with a population size of 250.
    \item Pareto Local Search (PLS) \cite{paquete_pareto_2004} is a very straightforward discrete multi-objective algorithm based on local search. PLS demonstrates great performance on the unconstrained multi-objective traveling salesman problem.
    \item MOEA/D \cite{zhang_moead_2007} uses decomposition and genetic operators to iteratively slove single-objective decompositions of a multi-objective problem, gradually reaching a Pareto front approximation. It is widely used in multi-objective optimization problems due to its computational efficiency and high performance.
    \item Pareto-MCTS \cite{chen_pareto_2019} is an adaptation of UCT to multi-objective optimization. The comparison of Pareto-NRPA to this work is relevant because both methods use random playouts to optimize a Pareto front. The two algorithms are related to Monte-Carlo Tree Search.
\end{itemize}

SMS-EMOA, MOEA/D and NSGA-II are benchmarked using their respective implementation in the PyMoo framework \cite{blank_pymoo_2020}. 

\subsection*{MO-TSPTW}

\begin{figure}
    \centering
    \includegraphics[width=0.9\linewidth]{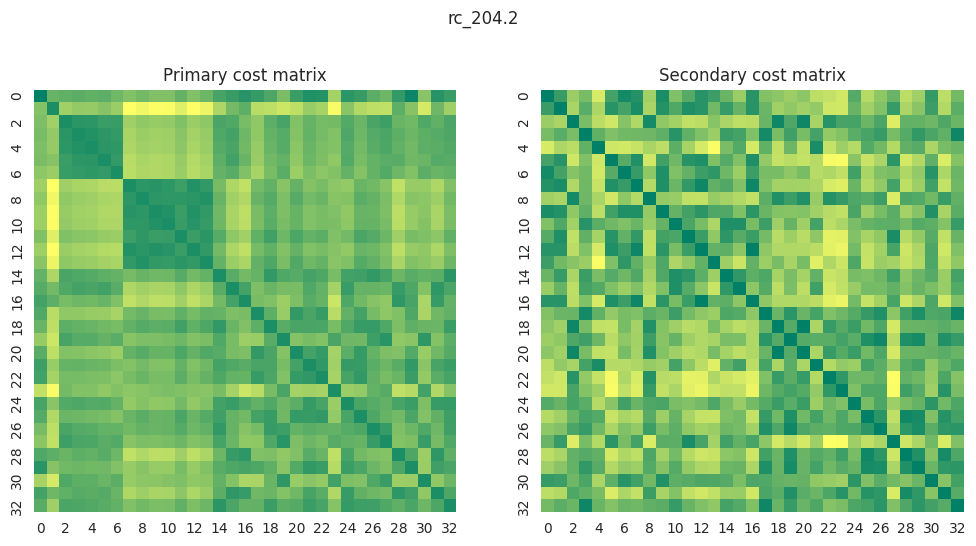}
    \caption{The two independent cost matrices for instance rc\_204.2 of the TSPTW}
    \label{fig:rc_204_2_cost}
\end{figure}

In this work, we introduce a novel bi-objective optimization dataset based on the TSPTW problem. In the classical version of the TSPTW problem, one aims to find a tour (permutation) of $n$ cities such that the tour starts and ends at city 0, each city is visited exactly once and each city is visited inside a particular time window. NRPA has shown promising results on this problem, finding state-of-the-art solutions on numerous instances \cite{cazenave_application_2012}. For each instance of the famous Solomon-Potvin-Bengio dataset \cite{potvin_vehicle_1996}, we generate a new set of coordinates for each city, and set the secondary cost associated to the travel from city $i$ to city $j$ as the euclidean distance between $i$ and $j$. Figure \ref{fig:rc_204_2_cost} shows the primary and secondary cost matrices for instance rc\_204.2 of the MO-TSPTW dataset. The time windows associated to each city are identical to the classical TSPTW instances. The primary and secondary distances between two cities are unrelated. As such, we find that minimizing both objectives at the same time is a challenging task, suitable for evaluating Pareto-NRPA. 
The multi-objective optimization problem associated with MO-TSPTW is: 
\begin{equation}
    \text{minimize } \begin{cases}
        f_1(s) = cost_1(s) + 10^6 \times \Omega(s) \\ 
        f_2(s) = cost_2(s) + 10^6 \times \Omega(s)
    \end{cases} 
\end{equation}
where : 
\begin{itemize}
    \item $cost_1(s)$ (resp. $cost_2(s)$) is the sum of the primary (resp. secondary) distances between each city and the city that follows it in the sequence $s$.
    \item $\Omega(s)$ is the number of violated constraints (i.e. the number of cities that have been visited outside of their respective time window).
\end{itemize}
The $10^6$ penalty applied to violated constraints quickly forces the algorithms to find valid sequences, as any valid solution, however inefficient it might be, always dominates an invalid solution.

The MO-TSPTW dataset contains 31 instances. For three of those instances, complete algorithm results (including hypervolume, overall spread, spacing and constraint violations) are presented and discussed: 
\begin{itemize}
    \item rc\_204.3 (24 cities) is the instance with the largest average time window per city. Thus, it is very easy to generate a valid solution that doesn't violate any constraints.
    \item rc\_201.3 (32 cities) is the instance with the narrowest average time window per city. It is particularly hard to generate a valid solution for this instance.
    \item rc\_204.1 (46 cities) is the instance with the largest number of cities to visit. While this instance has the second largest average time window in the dataset, the large number of cities strongly increases the complexity of the dataset: as such, satisfying the time window constraints for this instance is moderately hard.
\end{itemize}
Those three instances are representative of varying degrees of difficulty in the dataset and allow us to discuss different problem configurations. For all 31 instances of the dataset, the hypervolume indicator is presented in this section (Table \ref{table:full-tsptw-hv-part1}). Complete results and metrics, including hypervolume, overall spread, spacing and average constraint violations for all 31 instances are shown in the supplementary material (Section \ref{sec:supplementary}, Tables \ref{table:full-tsptw-hv}-\ref{table:full-tsptw-success}). These tables include 95\% confidence intervals : results for a metric $X$ are shown as $\bar{X} \pm CI$, where $\bar{X} = \frac{1}{n\_runs} \sum_{i=0}^{n\_runs} X_i$ is the average value of a metric over all runs, and $CI = 1.96 * \frac{\sigma}{\sqrt{n\_runs}}$, where $\sigma$ is the standard deviation of $X$. For space reasons, Table \ref{table:full-tsptw-hv-part1} omits the confidence intervals, which may be found in the supplementary material.

\begin{figure}[h]
  \centering
  \subfigure[]{\includegraphics[width=.48\linewidth]{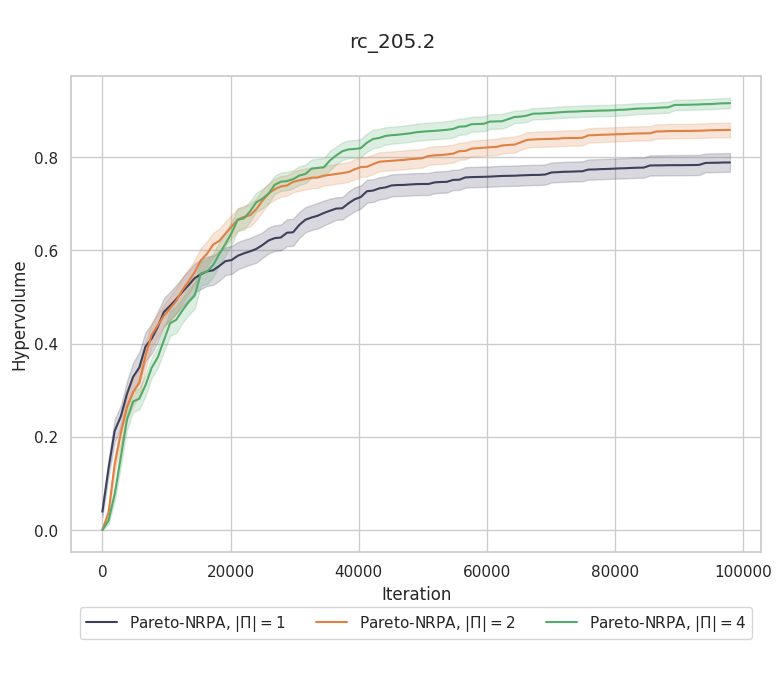}} \hfill
  \subfigure[]{\includegraphics[width=.48\linewidth]{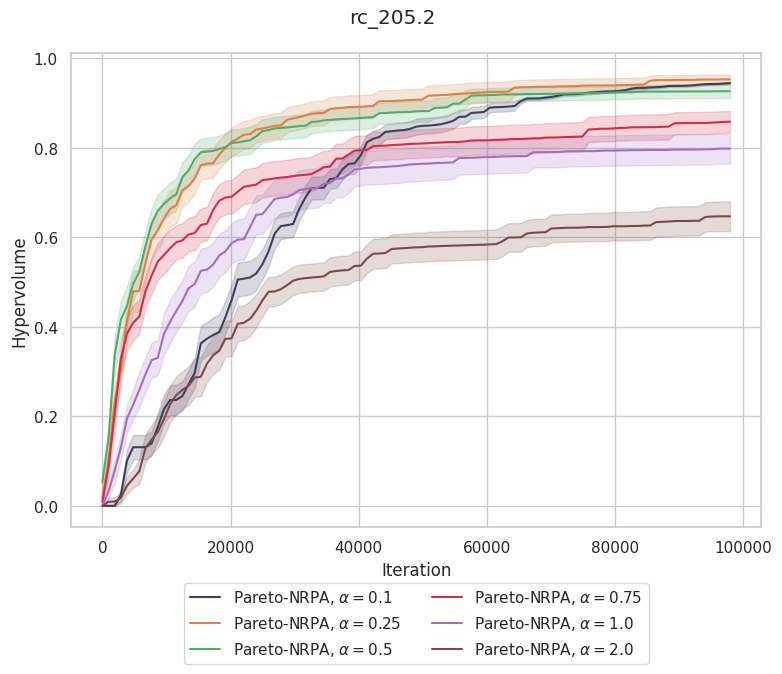}}
  \caption{Different values of $|\Pi|$ and $\alpha$ on rc\_205.2}  \label{fig:n-policies} 
\end{figure}

The run configurations of each algorithm are the following:
\begin{itemize}
    \item NSGA-II and SMS-EMOA: For both MOEAs, a member of the population is represented by a vector of size $n\_cities$. The sampling operator samples a permutation of all cities. The mutation operator randomly inverts the order of a random subsequence in the solution \cite{peng_traveling_2020}. Floating point values obtained after mutation and crossover are corrected using a rounding operation. The initial population size is 250.
    \item PLS: The initial population consists of 250 individuals.
    \item MOEA/D uses the same genetic parameters as NSGA-II and SMS-EMOA. The bi-objective problem is decomposed into 200 single-objective optimization subproblems.
    \item Pareto-MCTS: Pareto-MCTS implements the UCT algorithm \cite{kocsis_bandit_2006} with leaf parallelization for improved stability \cite{cazenave_parallelization_nodate} and speed. The UCB formula is parametrized by an exploration constant $C$. For optimization problems with rewards between 0 and 1, the exploration constant is often set to small values. However, as the results values of a playout in the two studied problem domains may be far greater than 1, we set $C$ as a dynamic value computed for every node $i$ : $C_i = \bar{r}_{\max} - \bar{r}_{\min}$, where $\bar{r}_{\max}$ (resp. $\bar{r}_{\min}$) is the maximal (resp. minimal) average node value for children of $i$.
    \item Pareto-NRPA: The same configuration of Pareto-NRPA is used for all instances in MO-TSPTW. The initial level $L$ is set as 4 and the learning rate $\alpha$ as $0.5$. The number of policies in $\Pi$ is 4. These hyperparameters have been chosen after a succinct grid search on the rc\_205.2 instance. Hypervolume evolution for different values of $|\Pi|$ and $\alpha$ are shown in Figures \ref{fig:n-policies}a and \ref{fig:n-policies}b. Complete hyperparameter search metrics for values of $\alpha \in \{0.1, 0.25, 0.5, 0.75, 1, 2 \}$ and $|\Pi| \in \{1, 2, 4 \}$ are presented in the supplementary material. Although tuning the hyperparameters for each instance would likely lead to improved performance, we intentionally use a fixed configuration to enable a fair comparison with other algorithms, which are also evaluated without instance-specific tuning. 
\end{itemize}

All algorithms are run for 100 000 objective function evaluations. 30 independent runs are performed for each algorithm.

It has been shown that adding a bias to the probabilities of action selection strongly improves NRPA performance \cite{cazenave_generalized_2020}. A bias value classically used for TSPTW is $-10*d_{ij}$, with $d_{ij}$ the normalized distance between two cities \cite{cazenave_generalized_2024}. For fairness concerns, we also implement the same bias term in the sampling operators for SMS-EMOA and NSGA-II, as well as in the playout algorithm in Pareto-MCTS. An ablation study done on SMS-EMOA and NSGA-II (Section \ref{sec:supplementary}, Table \ref{table:hv-bias-emoa}) confirms that the introduction of bias significantly improves MOEA performance, as the algorithms find better solutions during the sampling step.
In Tables \ref{table:metrics-rc204-3}, \ref{table:metrics-rc201-3} and \ref{table:metrics-rc204-1}, the hypervolume values are calculated with respect to a vector $r = (\max_{y} f_1(y), \max_y f_2(y))$ for all valid solution vectors $y$ in the aggregation of results of all 30 runs for all algorithms combined. The hypervolume indicator is then normalized using maximal hypervolume value after algorithm termination. Constraint violations (CV) refers to the average number of violated constraints for the best solution(s) over all 30 runs. The spacing metric is not computed for runs where no valid sequence has been found.

\begin{table}[t]
\footnotesize
\centering
\renewcommand{\arraystretch}{1.4}
\caption{Metrics on rc\_204.3}
\label{table:metrics-rc204-3}
\begin{tabular}{lcccr}
\toprule
Algorithm & Hypervolume & Overall Spread & Spacing & CV\\
\midrule
NSGA-II     & \bm{$0.97$} & $0.37$      & $5.33$      & \bm{$0.00$} \\
SMS-EMOA    & $0.96$      & $0.33$      & $7.06$      & \bm{$0.00$} \\
Pareto-MCTS & $0.61$      & $0.16$      & $18.08$     & \bm{$0.00$} \\
PLS         & $0.91$      & $0.37$      & \bm{$5.18$} & \bm{$0.00$} \\
MOEA/D      & $0.93$      & \bm{$0.38$} & $9.66$      & \bm{$0.00$} \\
Pareto-NRPA & $0.94$      & $0.32$      & $6.19$      & \bm{$0.00$} \\
\bottomrule
\end{tabular}
\end{table}

\begin{table}[H]
\centering
\renewcommand{\arraystretch}{1.4}
\caption{Metrics on rc\_201.3}
\label{table:metrics-rc201-3}
\begin{tabular}{lcccr}
\toprule
Algorithm & Hypervolume & Overall Spread & Spacing & CV \\
\midrule
NSGA-II     & $0.00$      & $0.00$      & -           & $7.33$      \\
SMS-EMOA    & $0.00$      & $0.00$      & -           & $5.97$      \\
Pareto-MCTS & $0.00$      & $0.00$      & -           & $9.63$      \\
PLS         & $0.00$      & $0.00$      & -           & $11.80$      \\
MOEA/D      & $0.00$      & $0.00$      & -           & $5.53$      \\
Pareto-NRPA & \bm{$0.91$} & \bm{$0.35$} & \bm{$4.38$} & \bm{$0.00$} \\ 
\bottomrule
\end{tabular}
\end{table}

\begin{table}[H]
\centering
\renewcommand{\arraystretch}{1.4}
\caption{Metrics on rc\_204.1}
\label{table:metrics-rc204-1}
\begin{tabular}{lcccr}
\toprule
Algorithm & Hypervolume & Overall Spread & Spacing & CV \\
\midrule
NSGA-II     & $0.12$      & $0.06$      & $12.30$      & $1.90$      \\
SMS-EMOA    & $0.00$      & $0.00$      & -            & $2.17$      \\
Pareto-MCTS & $0.00$      & $0.00$      & -            & $20.47$     \\
PLS         & $0.00$      & $0.00$      & -            & $5.67$      \\
MOEA/D      & $0.01$      & $0.07$      & \bm{$4.98$}       & $2.23$      \\
Pareto-NRPA & \bm{$0.26$} & \bm{$0.08$} & $10.12$ & \bm{$0.00$} \\ 
\bottomrule
\end{tabular}
\end{table}

\begin{table}[!t]
\tiny
\centering
\renewcommand{\arraystretch}{1.6}
\caption{Normalized hypervolume on all instances of MO-TSPTW}
\label{table:full-tsptw-hv-part1}

\begin{tabular}{lccccccr}
  \toprule
  Instance & Cities & NSGA-II & SMS-EMOA & Pareto-UCT & PLS & MOEA/D & Pareto-NRPA \\
  \midrule
  rc\_206.1 &  4 & \bm{$1.00$} & \bm{$1.00$} & \bm{$1.00$} & \bm{$1.00$} & \bm{$1.00$} & \bm{$1.00$}\\
  rc\_207.4 &  6 & \bm{$1.00$} & \bm{$1.00$} & \bm{$1.00$} & \bm{$1.00$} & \bm{$1.00$} & \bm{$1.00$}\\
  rc\_203.4 & 15 & \bm{$1.00$} & \bm{$1.00$} & $0.65$ & $0.97$ & $0.99$ & $0.95$\\
  rc\_202.2 & 14 & \bm{$1.00$} & \bm{$1.00$} & $0.84$ & $0.91$ & $0.99$ & $0.98$\\
  rc\_204.3 & 24 & \bm{$0.97$} & $0.96$ & $0.61$ & $0.91$ & $0.93$ & $0.94$\\
  rc\_203.1 & 19 & $0.91$ & $0.87$ & $0.44$ & $0.72$ & $0.88$ & \bm{$0.97$}\\
  rc\_204.2 & 33 & $0.80$ & $0.79$ & $0.00$ & $0.80$ & $0.57$ & \bm{$0.82$} \\
  rc\_208.2 & 29 & \bm{$0.94$} & $0.93$ & $0.00$ & $0.89$ & $0.84$ & $0.86$\\
  rc\_204.4 & 14 & \bm{$1.00$} & \bm{$1.00$} & $0.60$ & $0.89$ & $0.99$ & $0.99$ \\
  rc\_205.1 & 14 & \bm{$1.00$}  & \bm{$1.00$} & $0.63$ & $0.98$ & $0.99$ & \bm{$1.00$} \\
  rc\_204.1 & 46 & $0.12$ & $0.00$ & $0.00$ & $0.00$ & $0.01$ & \bm{$0.29$}\\
  rc\_203.2 & 33 & $0.11$ & $0.27$ & $0.00$ & $0.05$ & $0.21$ & \bm{$0.86$} \\
  rc\_203.3 & 37 & $0.08$ & $0.04$ & $0.00$ & $0.03$ & $0.04$ & \bm{$0.90$} \\
  rc\_208.3 & 36 & $0.92$ & \bm{$0.93$} & $0.00$ & $0.57$ & $0.63$ & $0.81$ \\
  rc\_202.4 & 28 & $0.01$ & $0.07$ & $0.00$ & $0.00$ & $0.00$ & \bm{$0.77$}\\
  rc\_208.1 & 38 & $0.06$ & $0.06$ & $0.00$ & $0.02$ & $0.01$ & \bm{$0.46$}\\
  rc\_207.3 & 33 & $0.24$ & $0.28$ & $0.00$ & $0.05$ & $0.37$ & \bm{$0.92$}\\
  rc\_207.2 & 31 & $0.00$ & $0.08$ & $0.00$ & $0.00$ & $0.02$ & \bm{$0.33$}\\
  rc\_206.3 & 25 & $0.56$ & $0.63$ & $0.01$ & $0.03$ & $0.79$ & \bm{$0.94$}\\
  rc\_207.1 & 34 & $0.42$ & $0.44$ & $0.00$ & $0.00$ & $0.34$ & \bm{$0.79$}\\
  rc\_202.1 & 33 & $0.00$ & $0.08$ & $0.00$ & $0.02$ & $0.04$ & \bm{$0.91$}\\
  rc\_205.3 & 35 & $0.27$ & $0.25$ & $0.00$ & $0.00$ & $0.00$ & \bm{$0.77$}\\  
  rc\_201.1 & 20 & $0.77$ & $0.73$ & $0.61$ & $0.28$ & $0.93$ & \bm{$0.97$}\\
  rc\_205.2 & 27 & $0.00$ & $0.00$ & $0.00$ & $0.00$ & $0.01$ & \bm{$0.92$} \\
  rc\_205.4 & 28 & $0.25$ & $0.28$ & $0.00$ & $0.00$ & $0.22$ & \bm{$0.80$} \\
  rc\_202.3 & 29 & $0.00$ & $0.00$ & $0.00$ & $0.00$ & $0.00$ & \bm{$0.94$}\\
  rc\_206.2 & 37 & $0.00$ & $0.00$ & $0.00$ & $0.00$ & $0.00$ & \bm{$0.76$} \\
  rc\_206.4 & 38 & $0.00$ & $0.00$ & $0.00$ & $0.00$ & $0.00$ & \bm{$0.51$}\\
  rc\_201.2 & 26 & $0.06$ & $0.25$ & $0.00$ & $0.00$ & $0.14$ & \bm{$0.84$}\\
  rc\_201.4 & 26 & $0.00$ & $0.08$ & $0.00$ & $0.00$ & $0.03$ & \bm{$0.33$}\\
  rc\_201.3 & 32 & $0.00$ & $0.00$ & $0.00$ & $0.00$ & $0.00$ & \bm{$0.93$}\\

  \bottomrule
\end{tabular}

\end{table}

Table \ref{table:metrics-rc204-3} indicates that NSGA-II and SMS-EMOA converge to a slightly better solution set than Pareto-NRPA on a very easy instance. Every algorithm succeeds in finding valid solutions for all 30 runs. However, Tables \ref{table:metrics-rc201-3} and \ref{table:metrics-rc204-1} show that, on the two harder instances, Pareto-NRPA converges to the best solutions, illustrated by the superior hypervolume values. The evolutionary algorithms, PLS and Pareto-MCTS fail to produce valid solutions for almost every run, while Pareto-NRPA converges quickly to solutions that do not violate time constraints. A hypervolume value of $0$ indicates that the solution set generated by the algorithm is dominated by the reference point, which means that no valid solution has been found.

The results in Table \ref{table:full-tsptw-hv-part1} show that Pareto-NRPA strongly outperforms the state-of-the-art evolutionary MOO algorithms on most instances of the multi-objective TSPTW dataset. Precisely, Pareto-NRPA returns a greatly superior hypervolume metric to the other state-of-the-art algorithms on 22 out of the 31 problem instances of MO-TSPTW. Evolutionary algorithms match or slightly outperform (in 5 cases) Pareto-NRPA in 8 of the 10 easiest instances in the dataset, while Pareto-NRPA is the superior algorithm in 20 of the 21 remaining hardest instances of the dataset. We note that Pareto-MCTS performs poorly on most hard instances of the dataset. Indeed, the UCB formula \cite{kocsis_bandit_2006} is not suited to constraint handling. Setting the reward with a large penalty for violated constraints prevents efficient node value estimation, and simply aborting a playout where constraints are violated often leads to suboptimal, over-conservative policies \cite{lee_monte-carlo_2018}. While Pareto-NRPA and Pareto-MCTS are both based on Monte-Carlo playouts, these results show that Pareto-NRPA is much more suited to constrained search spaces.

NSGA-II, MOEA/D and SMS-EMOA converge to good and diverse solution sets on easy instances (such as rc\_202.2 or rc\_205.1, even slightly outperforming Pareto-NRPA), but generally fail to converge to a valid solution for harder instances, while Pareto-NRPA succeeds in finding solutions that don't violate any time constraints. Indeed, the policy adaptation mechanism implemented in Pareto-NRPA allows the algorithm to learn sequences of moves which leads to faster convergence. Pareto-NRPA thus emerges as an efficient algorithm for constraint handling in sequential discrete multi-objective optimization problems, strongly outperforming MOEAs on problems with constraints that are hard to satisfy.

We note that Pareto Local Search, despite being very popular on the unconstrained multi-objective TSP, doesn't yield good results on MO-TSPTW. This can be explained by the fact that the algorithms are capped at 100 000 function evaluations, a relatively low value for PLS which requires more iterations to efficiently evaluate diverse regions of the search space. Further experiments comparing the algorithms under equal search times show that PLS reaches good performances with more function evaluations (Section \ref{sec:supplementary}, Tables \ref{table:metrics-rc204-3-equal-cpu}-\ref{table:metrics-rc204-1-equal-cpu}).

\begin{figure}[h]
  \centering
  \subfigure[]{\includegraphics[width=.48\linewidth]{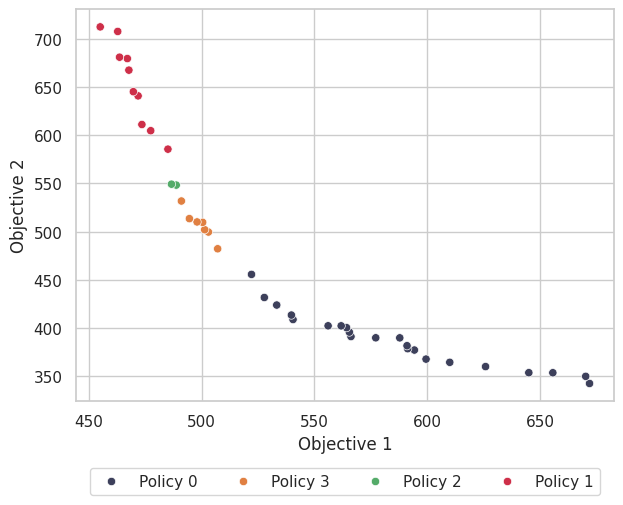}} \hfill
  \subfigure[]{\includegraphics[width=.48\linewidth]{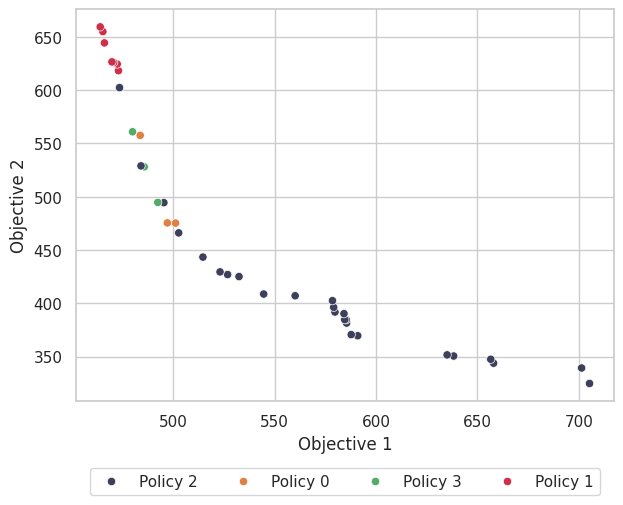}}
  \caption{Policy distribution for two different runs on rc\_204.3. Each color represents the policy from which the solution has been sampled.}  \label{fig:policy-distribution} 
\end{figure}

The diversity of the solutions produced by Pareto-NRPA is not always as good as the MOEAs. This is due to the fact that MOEAs maintain a large population, which retains diversity, while Pareto-NRPA only maintains a front of non-dominated sequences. Pareto-NRPA does not explicitely enforce diversity of solutions and their associated policy weights: the policy update is merely performed with respect to the crowding distance of sequences in $S^*$. As such, the overall distribution of policies in the Pareto front approximation may be uneven. Figure \ref{fig:policy-distribution} shows the distribution of policies over the generated solutions for two different runs on rc\_204.3. It may be appreciated that one run has converged to policies focusing on different regions of the search space, leading to varied solutions, while the other has a less distinct separation of policies in the solution space. Improving Pareto-NRPA's distribution of policies over different regions of the search space remains an axis for future work.

It is worth noting that Pareto-NRPA is computationally heavier than the compared state-of-the-art algorithms, especially the extremely fast MOEA/D and PLS. When the size of the search space increases, the complexity of Pareto-NRPA grows. Experimental results comparing the algorithms under equal CPU time are shown in the supplementary material of this paper, and demonstrate that Pareto-NRPA can be outperformed under strict search time conditions. Improving the computational complexity of Pareto-NRPA is a very clear research avenue.

\FloatBarrier

\subsection*{Neural Architecture Search}

Neural Architecture Search (NAS) refers to the task of automatically exploring a search space to find a neural network architecture minimizing one, or several, metrics. A classical goal for single-objective NAS is finding the architecture according to: 
\begin{equation}
    a^* = \arg \min_{a \in S} \mathcal{L}(X, Y, W_a)
\end{equation}
 where $S$ is the space of all architectures, $W_a$ are the neural network weights associated to architecture $a$ and $\mathcal{L}(X, Y, W_a)$ is the loss computed on the validation dataset. The search space $S$ is, in practical applications, typically large ($|S| > 10^{10}$). 
Recently, tabular benchmarks have been proposed to facilitate NAS research. These tabular benchmarks contain training and validation accuracies for every neural network in a restrained search space. Some of these benchmarks are NAS-Bench-201 \cite{dong_nas-bench-201_2020}, which contains 15625 architectures, and NAS-Bench-101 \cite{ying_nas-bench-101_2019}, with 423 000 neural network architectures. We use these two benchmarks to quickly evaluate the metrics associated to neural networks in a multi-objective setting. The two objective functions we aim to optimize are :
\begin{itemize}
    \item $f_1 = 100 - Acc(a)$: the classification error of network $a$ computed over the CIFAR-10 validation set.
    \item $f_2 = \#params$: the number of parameters of the neural network.
\end{itemize}
Minimizing these two metrics simultaneously amounts to finding an efficient trade-off between validation accuracy and neural network computational complexity. Indeed, it may be of interest to deep learning practicioners to identify high-performing neural networks with a limited resource budget in terms of inference speed of memory constraints. Multi-objective NAS approaches have been proposed using evolutionary algorithms \cite{elsken_efficient_2019} \cite{lu_nsga-net_2019} and reinforcement learning \cite{hsu_monas_2018}. Moreover, we intentionally limit the number of objective function evaluation to a low value (2000 function evaluations) as a single function evaluation is often very costly in NAS \cite{zoph_neural_2016}.

The first NAS benchmark dataset is NAS-Bench-201 \cite{dong_nas-bench-201_2020}. It is a cell-based search space where a cell is represented as a directed acyclic graph with 4 vertices. Each edge $(u, v)$ in the DAG is associated with one of the following operations applied to vertex $u$, transforming it to vertex $v$: skip-connection, 1x1 convolution, 3x3 convolution, 3x3 average pooling, or no operation at all. Searching a cell thus consists in assigning one of the available operations to each edge. The searched cell is finally used to create a neural network according to a pre-defined skeleton. We refer the reader to \cite{dong_nas-bench-201_2020} for more information concering NAS-Bench-201.

\begin{figure}[h]
  \centering
  \subfigure[]{\includegraphics[width=.48\linewidth]{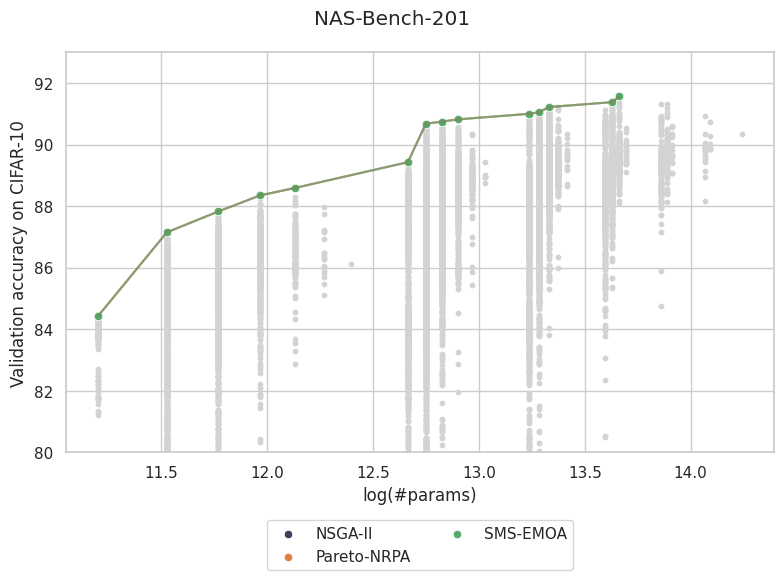}} \hfill
  \subfigure[]{\includegraphics[width=.48\linewidth]{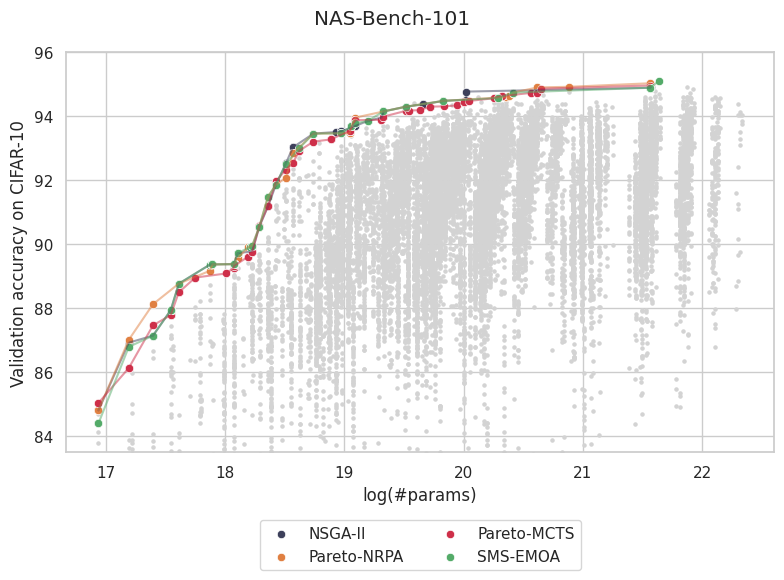}}
  \caption{Aggregated Pareto fronts on NAS-Bench-201 and NAS-Bench-101}  \label{fig:nas} 
\end{figure}

NSGA-II, SMS-EMOA, Pareto-MCTS and Pareto-NRPA are independently run 30 times on the NAS-Bench-201 dataset. Each run consists in 2000 objective function evaluations. 
Figure \ref{fig:nas}a plots the accuracy of the neural networks found during search ($100 - f_1)$. It indicates that all algorithms have succeeded in finding the true Pareto front for NAS-Bench-201 after 30 runs. The Pareto fronts are superposed, meaning that all the algorithms have found the same set of points. Data points plotted in grey refer to the available architectures in the search space. It is expected that all algorithms succeed in finding the optimal Pareto front since the search space is small ($|S| = 15625$). Computing the average overall spread and hypervolume metrics however indicates that some runs converge to slightly suboptimal Pareto front approximations (Table \ref{table:metrics-nb201}). While all algorithms show very good performance, NSGA-II and SMS-EMOA obtain a hypervolume value 0.01 larger than Pareto-NRPA.

\begin{table}
\centering
\renewcommand{\arraystretch}{1.4}
\caption{Metrics on NAS-Bench-201}
\label{table:metrics-nb201}
\begin{tabular}{lcccr}
\toprule
Algorithm & Hypervolume & Overall Spread \\
\midrule
NSGA-II     & \bm{$0.99 \pm 0.00$}      & \bm{$0.72 \pm 0.02$}      \\
SMS-EMOA    & \bm{$0.99 \pm 0.00$}     & $0.70 \pm 0.03$      \\
Pareto-MCTS & $0.96 \pm 0.00$     & $0.68 \pm 0.05$     \\
Pareto-NRPA & $0.98 \pm 0.00$ & $0.70 \pm 0.03$ \\ 
\bottomrule
\end{tabular}

\end{table}

\FloatBarrier

The second NAS benchmark dataset is NAS-Bench-101 \cite{ying_nas-bench-101_2019}. NAS-Bench-101 is also a cell-based search space, where each cell is encoded as a DAG with 7 vertices. While NAS-Bench-201 associates operations to graph edges and feature maps to vertices, NAS-Bench-101 associates operations to vertices. The available operations are 1x1 convolution, 3x3 convolution and 3x3 max pooling. There are a total of 423 000 valid architectures in the NAS-Bench-101 search space: as such, finding a good Pareto front approximation on this dataset is slightly harder than on NAS-Bench-201. 
Each algorithm is run for 2000 objective function evaluations for 30 runs. 

\begin{table}
\centering
\renewcommand{\arraystretch}{1.4}
\caption{Metrics on NAS-Bench-101}
\label{table:metrics-nb101}
\begin{tabular}{lcccr}
\toprule
Algorithm & Hypervolume & Overall Spread \\
\midrule
NSGA-II     & $0.98 \pm 0.00$      & $0.64 \pm 0.04$      \\
SMS-EMOA    & $0.97 \pm 0.00$      & $0.62 \pm 0.03$      \\
Pareto-MCTS & $0.97 \pm 0.00$     &  $0.56 \pm 0.03$    \\
Pareto-NRPA & \bm{$0.99 \pm 0.00$} & \bm{$0.72 \pm 0.04$} \\ 
\bottomrule
\end{tabular}

\end{table}

Results in Table \ref{table:metrics-nb101} and Figure \ref{fig:nas}b show that, once again, all algorithms succeed in finding good solutions sets for NAS-Bench-101. This time, Pareto-NRPA slightly outperforms the state-of-the-art MOEAs. The two NAS search spaces presented are both unconstrained, which means that no neural network architecture is considered invalid. As such, MOEAs are well-suited to the task and Pareto-NRPA does not outperform them as strongly as in constrained search spaces such as harder instances of MO-TSPTW.

\FloatBarrier

\section{Conclusion}

We have proposed Pareto-NRPA, a novel multi-objective Monte-Carlo search algorithm using nested rollouts and policy adaptation. Pareto-NRPA uses a set of policies to guide a recursive search over several nested levels, and adapts the policy weights based on non-dominated solutions in the objective space. We benchmark Pareto-NRPA on a novel dataset extending classical instances of the Traveling Salesman Problem with Time Windows to multi-objective optimization (MO-TSPTW), as well as a neural architecture search task (NAS). We report that Pareto-NRPA outperforms state-of-the-art multi-objective evolutionary algorithms on MO-TSPTW and leads to competitive results on the NAS task. Pareto-NRPA is identified as a strong algorithm for constraint handling in sequential discrete multi-objective optimization problems. Future research directions include:
\begin{itemize}
    \item Extending Pareto-NRPA to continuous search spaces.
    \item Benchmarking Pareto-NRPA on multi-objective optimization problems with 3 and more objectives.
    \item Improving policy diversity in the solution set.
    \item Accelerating Pareto-NRPA by reducing the computational complexity of the algorithm.
\end{itemize}





\bibliography{mybibfile}

\newpage

\onecolumn

\section{Supplementary material} \label{sec:supplementary}

\subsection{Full results on MO-TSPTW}

The full results on TSPTW are shown in Tables \ref{table:full-tsptw-hv}, \ref{table:full-tsptw-spacing}, \ref{table:full-tsptw-spread} and \ref{table:full-tsptw-success}. Specifically, 
\begin{itemize}
    \item Table \ref{table:full-tsptw-hv} shows hypervolumes for all instances of MO-TSPTW
    \item Table \ref{table:full-tsptw-spread} shows the overall spread (OS) metric
    \item Table \ref{table:full-tsptw-spacing} shows the spacing metric
    \item Table \ref{table:full-tsptw-success} shows the average constraint violations
\end{itemize}
These tables include 95\% confidence intervals : results for a metric $X$ are shown as $\bar{X} \pm CI$, where $\bar{X} = \frac{1}{n\_runs} \sum_{i=0}^{n\_runs} X_i$ is the average value of a metric over all runs, and $CI = 1.96 * \frac{\sigma}{\sqrt{n\_runs}}$, where $\sigma$ is the standard deviation of $X$. 
The instances are ordered by ascending ratio between average time window width of cities and number of cities. While not exactly representative of instance hardness, this ordering gives an idea of the relative ease of satisfying the constraints.

\begin{table}
\centering
\renewcommand{\arraystretch}{1.6
}

\begin{tabular}{lccccr}
  \toprule
  Instance & Cities & NSGA-II & SMS-EMOA & Pareto-UCT & Pareto-NRPA \\
  \midrule
  rc\_206.1 &  4 & \bm{$1.00 \pm 0.00$} & \bm{$1.00 \pm 0.00$} & \bm{$1.00 \pm 0.00$} & \bm{$1.00 \pm 0.00$}\\
  rc\_207.4 &  6 & \bm{$1.00 \pm 0.00$} & \bm{$1.00 \pm 0.00$} & \bm{$1.00 \pm 0.00$} & \bm{$1.00 \pm 0.00$}\\
  rc\_203.4 & 15 & \bm{$1.00 \pm 0.00$} & \bm{$1.00 \pm 0.00$} & $0.65 \pm 0.01$ & $0.95 \pm 0.01$\\
  rc\_202.2 & 14 & \bm{$1.00 \pm 0.00$} & \bm{$1.00 \pm 0.00$} & $0.84 \pm 0.01$ & $0.98 \pm 0.00$\\
  rc\_204.3 & 24 & \bm{$0.97 \pm 0.01$} & $0.96 \pm 0.01$ & $0.61 \pm 0.02$ & $0.94 \pm 0.01$\\
  rc\_203.1 & 19 & $0.91 \pm 0.02$ & $0.87 \pm 0.03$ & $0.44 \pm 0.04$ & \bm{$0.97 \pm 0.01$}\\
  rc\_204.2 & 33 & $0.80 \pm 0.03$ & $0.79 \pm 0.04$ & $0.00 \pm 0.00$ & \bm{$0.82 \pm 0.02$} \\
  rc\_208.2 & 29 & \bm{$0.94 \pm 0.01$} & $0.93 \pm 0.01$ & $0.00 \pm 0.00$ & $0.86 \pm 0.02$\\
  rc\_204.4 & 14 & \bm{$1.00 \pm 0.00$} & \bm{$1.00 \pm 0.00$} & $0.60 \pm 0.04$ & $0.99 \pm 0.01$ \\
  rc\_205.1 & 14 & \bm{$1.00 \pm 0.00$}  & \bm{$1.00 \pm 0.00$} & $0.63 \pm 0.04$ & \bm{$1.00 \pm 0.01$} \\
  rc\_204.1 & 46 & $0.12 \pm 0.11$ & $0.00 \pm 0.00$ & $0.00 \pm 0.00$ & \bm{$0.29 \pm 0.05$}\\
  rc\_203.2 & 33 & $0.11 \pm 0.08$ & $0.27 \pm 0.12$ & $0.00 \pm 0.00$ & \bm{$0.86 \pm 0.02$} \\
  rc\_203.3 & 37 & $0.08 \pm 0.07$ & $0.04 \pm 0.05$ & $0.00 \pm 0.00$ & \bm{$0.90 \pm 0.02$} \\
  rc\_208.3 & 36 & $0.92 \pm 0.06$ & \bm{$0.93 \pm 0.01$} & $0.00 \pm 0.00$ & $0.81 \pm 0.02$ \\
  rc\_202.4 & 28 & $0.01 \pm 0.02$ & $0.07 \pm 0.06$ & $0.00 \pm 0.00$ & \bm{$0.77 \pm 0.08$}\\
  rc\_208.1 & 38 & $0.06 \pm 0.07$ & $0.06 \pm 0.07$ & $0.00 \pm 0.00$ & \bm{$0.46 \pm 0.06$}\\
  rc\_207.3 & 33 & $0.24 \pm 0.12$ & $0.28 \pm 0.14$ & $0.00 \pm 0.00$ & \bm{$0.92 \pm 0.02$}\\
  rc\_207.2 & 31 & $0.00 \pm 0.00$ & $0.08 \pm 0.09$ & $0.00 \pm 0.00$ & \bm{$0.33 \pm 0.11$}\\
  rc\_206.3 & 25 & $0.56 \pm 0.17$ & $0.63 \pm 0.16$ & $0.01 \pm 0.01$ & \bm{$0.94 \pm 0.02$}\\
  rc\_207.1 & 34 & $0.42 \pm 0.15$ & $0.44 \pm 0.14$ & $0.00 \pm 0.00$ & \bm{$0.79 \pm 0.03$}\\
  rc\_202.1 & 33 & $0.00 \pm 0.00$ & $0.08 \pm 0.09$ & $0.00 \pm 0.00$ & \bm{$0.91 \pm 0.02$}\\
  rc\_205.3 & 35 & $0.27 \pm 0.14$ & $0.25 \pm 0.13$ & $0.00 \pm 0.00$ & \bm{$0.77 \pm 0.05$}\\
  rc\_201.1 & 20 & $0.77 \pm 0.15$ & $0.73 \pm 0.16$ & $0.61 \pm 0.05$ & \bm{$0.97 \pm 0.00$}\\
  rc\_205.2 & 27 & $0.00 \pm 0.00$ & $0.00 \pm 0.00$ & $0.00 \pm 0.00$ & \bm{$0.92 \pm 0.05$} \\
  rc\_205.4 & 28 & $0.25 \pm 0.13$ & $0.28 \pm 0.14$ & $0.00 \pm 0.00$ & \bm{$0.80 \pm 0.03$} \\
  rc\_202.3 & 29 & $0.00 \pm 0.00$ & $0.00 \pm 0.00$ & $0.00 \pm 0.00$ & \bm{$0.94 \pm 0.01$}\\
  rc\_206.2 & 37 & $0.00 \pm 0.00$ & $0.00 \pm 0.00$ & $0.00 \pm 0.00$ & \bm{$0.76 \pm 0.07$} \\
  rc\_206.4 & 38 & $0.00 \pm 0.00$ & $0.00 \pm 0.00$ & $0.00 \pm 0.00$ & \bm{$0.51 \pm 0.08$}\\
  rc\_201.2 & 26 & $0.06 \pm 0.08$ &  $0.25 \pm 0.15$& $0.00 \pm 0.00$ & \bm{$0.84 \pm 0.03$}\\
  rc\_201.4 & 26 & $0.00 \pm 0.00$ & $0.08 \pm 0.09$ & $0.00 \pm 0.00$ & \bm{$0.33 \pm 0.11$}\\
  rc\_201.3 & 32 & $0.00 \pm 0.00$ & $0.00 \pm 0.00$ & $0.00 \pm 0.00$ & \bm{$0.93 \pm 0.02$}\\

  \bottomrule
\end{tabular}

\caption{Normalized hypervolume on all instances of MO-TSPTW}
\label{table:full-tsptw-hv}
\end{table}

\begin{table}
\centering
\renewcommand{\arraystretch}{1.6}
\caption{Pareto front spread on all instances of MO-TSPTW}
\label{table:full-tsptw-spread}
\begin{tabular}{lccccr}
  \toprule
  Instance & Cities & NSGA-II & SMS-EMOA & Pareto-MCTS & Pareto-NRPA \\
  \midrule
  rc\_206.1 &  4 & \bm{$1.00 \pm 0.00$} & \bm{$1.00 \pm 0.00$} & \bm{$1.00 \pm 0.00$} & \bm{$1.00 \pm 0.00$}\\
  rc\_207.4 &  6 & \bm{$1.00 \pm 0.00$} & \bm{$1.00 \pm 0.00$} & \bm{$1.00 \pm 0.00$} & \bm{$1.00 \pm 0.00$}\\
  rc\_203.4 & 15 & \bm{$0.58 \pm 0.01$} & $0.58 \pm 0.03$ & $0.29 \pm 0.07$ & $0.30 \pm 0.05$\\
  rc\_202.2 & 14 & \bm{$0.79 \pm 0.00$} & \bm{$0.79 \pm 0.00$} & $0.40 \pm 0.04$ & $0.26 \pm 0.03$\\
  rc\_204.3 & 24 & \bm{$0.37 \pm 0.03$} & $0.33 \pm 0.02$ & $0.16 \pm 0.02$ & $0.32 \pm 0.02$\\
  rc\_203.1 & 19 & $0.28 \pm 0.04$ & $0.28 \pm 0.03$ & $0.05 \pm 0.02$ & \bm{$0.35 \pm 0.03$}\\
  rc\_204.2 & 33 & $0.41 \pm 0.04$ & $0.35 \pm 0.04$ & $0.00 \pm 0.00$ & \bm{$0.44 \pm 0.04$}\\
  rc\_208.2 & 29 & \bm{$0.40 \pm 0.04$} & $0.25 \pm 0.03$ & $0.00 \pm 0.00$ & $0.25 \pm 0.03$\\
  rc\_204.4 & 14 & \bm{$0.55 \pm 0.00$} & \bm{$0.55 \pm 0.00$} & $0.34 \pm 0.03$ & $0.54 \pm 0.01$\\
  rc\_205.1 & 14 & $0.70 \pm 0.02$ & \bm{$0.71 \pm 0.00$} & $0.35 \pm 0.04$ &  $0.67 \pm 0.03$\\
  rc\_204.1 & 46 & $0.06 \pm 0.06$ & $0.00 \pm 0.00$ & $0.00 \pm 0.00$ & \bm{$0.10 \pm 0.03$}\\
  rc\_203.2 & 33 & $0.03 \pm 0.03$ & $0.09 \pm 0.04$ & $0.00 \pm 0.00$ & \bm{$0.52 \pm 0.04$}\\
  rc\_203.3 & 37 & $0.04 \pm 0.03$ & $0.03 \pm 0.03$ & $0.00 \pm 0.00$ & \bm{$0.37 \pm 0.04$}\\
  rc\_208.3 & 36 & \bm{$0.37 \pm 0.05$} & $0.29 \pm 0.03$ & $0.00 \pm 0.00$ & $0.23 \pm 0.04$\\
  rc\_202.4 & 28 & $0.00 \pm 0.01$ & $0.02 \pm 0.02$ & $0.00 \pm 0.00$ & \bm{$0.40 \pm 0.06$}\\
  rc\_208.1 & 38 & $0.02 \pm 0.02$ & $0.02 \pm 0.02$ & $0.00 \pm 0.00$ & \bm{$0.13 \pm 0.04$}\\
  rc\_207.3 & 33 & $0.11 \pm 0.06$ & $0.10 \pm 0.05$ & $0.00 \pm 0.00$ & \bm{$0.44 \pm 0.05$}\\
  rc\_207.2 & 31 & $0.00 \pm 0.00$ & $0.03 \pm 0.03$ & $0.00 \pm 0.00$ & \bm{$0.10 \pm 0.04$}\\
  rc\_206.3 & 25 & $0.24 \pm 0.08$ & $0.27 \pm 0.09$ & $0.00 \pm 0.00$ & \bm{$0.44 \pm 0.05$}\\
  rc\_207.1 & 34 & $0.18 \pm 0.07$ & $0.22 \pm 0.08$ & $0.00 \pm 0.00$ & \bm{$0.39 \pm 0.04$}\\
  rc\_202.1 & 33 & $0.00 \pm 0.00$ & $0.03 \pm 0.04$ & $0.00 \pm 0.00$ & \bm{$0.45 \pm 0.06$}\\
  rc\_205.3 & 35 & $0.07 \pm 0.04$ & $0.12 \pm 0.07$ & $0.00 \pm 0.00$ & \bm{$0.16 \pm 0.04$}\\
  rc\_201.1 & 20 & \bm{$0.63 \pm 0.12$} & $0.60 \pm 0.13$ & $0.21 \pm 0.03$ & $0.42 \pm 0.06$\\
  rc\_205.2 & 27 & $0.00 \pm 0.00$ & $0.00 \pm 0.00$ & $0.00 \pm 0.00$ & \bm{$0.65 \pm 0.06$}\\
  rc\_205.4 & 28 & $0.11 \pm 0.07$ & $0.06 \pm 0.03$ & $0.00 \pm 0.00$ & \bm{$0.45 \pm 0.04$}\\
  rc\_202.3 & 29 & $0.00 \pm 0.00$ & $0.00 \pm 0.00$ & $0.00 \pm 0.00$ & \bm{$0.71 \pm 0.06$}\\
  rc\_206.2 & 37 & $0.00 \pm 0.00$ & $0.00 \pm 0.00$ & $0.00 \pm 0.00$ & \bm{$0.23 \pm 0.06$}\\
  rc\_206.4 & 38 & $0.00 \pm 0.00$ & $0.00 \pm 0.00$ & $0.00 \pm 0.00$ & \bm{$0.12 \pm 0.03$}\\
  rc\_201.2 & 26 &  $0.05 \pm 0.07$&  $0.17 \pm 0.10$& $0.00 \pm 0.00$ & \bm{$0.44 \pm 0.08$}\\
  rc\_201.4 & 26 & $0.00 \pm 0.00$ & $0.04 \pm 0.06$ & $0.00 \pm 0.00$ & \bm{$0.48 \pm 0.02$}\\
  rc\_201.3 & 32 & $0.00 \pm 0.00$ & $0.00 \pm 0.00$ & $0.00 \pm 0.00$ & \bm{$0.35 \pm 0.03$}\\
  \bottomrule
\end{tabular}

\end{table}

\begin{table}
\centering
\renewcommand{\arraystretch}{1.6}
\caption{Spacing on all instances of MO-TSPTW (lower is better)}
\label{table:full-tsptw-spacing}
\begin{tabular}{lccccr}
  \toprule
  Instance & Cities & NSGA-II & SMS-EMOA & Pareto-MCTS & Pareto-NRPA \\
  \midrule
  rc\_206.1 &  4 & \bm{$0.00 \pm 0.00$} & \bm{$0.00 \pm 0.00$} & \bm{$0.00 \pm 0.00$} & \bm{$0.00 \pm 0.00$}\\
  rc\_207.4 &  6 & \bm{$6.01 \pm 0.00$} & \bm{$6.01 \pm 0.00$} & \bm{$6.01 \pm 0.00$} & \bm{$6.01 \pm 0.00$}\\
  rc\_203.4 & 15 & $7.32 \pm 0.18$ & $7.51 \pm 0.42$ & $19.07 \pm 2.48$ & \bm{$6.17 \pm 0.92$}\\
  rc\_202.2 & 14 & \bm{$11.91 \pm 0.01$} & $11.96 \pm 0.07$ & $15.62 \pm 2.36$ & $13.56 \pm 0.13$ \\
  rc\_204.3 & 24 & \bm{$5.33 \pm 0.61$} & $7.06 \pm 1.19$ & $18.08 \pm 3.26$ & $6.19 \pm 0.56$\\
  rc\_203.1 & 19 & $7.34 \pm 1.07$ & \bm{$7.28 \pm 1.39$} & $15.67 \pm 7.11$ & $8.35 \pm 0.90$ \\
  rc\_204.2 & 33 & $11.54 \pm 1.86$ & $14.69 \pm 2.71$ & - & \bm{$10.43 \pm 1.21$} \\
  rc\_208.2 & 29 & \bm{$7.90 \pm 1.06$} & $10.26 \pm 2.63$ & - & $10.36 \pm 1.15$\\
  rc\_204.4 & 14 & \bm{$7.23 \pm 0.16$} & $7.28 \pm 0.07$ & $9.26 \pm 1.61$ & $7.25 \pm 0.24$ \\
  rc\_205.1 & 14 & $8.04 \pm 0.12$ & $8.02 \pm 0.2$ & $9.13 \pm 2.26$ & \bm{$7.75 \pm 0.26$} \\
  rc\_204.1 & 46 & $12.30 \pm 2.54$ & - & - & \bm{$10.68 \pm 2.20$}\\
  rc\_203.2 & 33 & $11.80 \pm 7.34$ & $9.74 \pm 5.21$ & - & \bm{$7.13 \pm 1.12$} \\
  rc\_203.3 & 37 & $8.21 \pm 2.19$ & $7.04 \pm 2.43$ & - & \bm{$4.51 \pm 0.74$} \\
  rc\_208.3 & 36 & \bm{$10.61 \pm 1.61$} & $11.13 \pm 1.66$ & - & $11.29 \pm 1.74$\\
  rc\_202.4 & 28 & $15.34 \pm 21.26$ & $12.20 \pm 5.45$ & - & \bm{$11.10 \pm 1.81$} \\
  rc\_208.1 & 38 & $11.92 \pm 5.45$ & \bm{$8.45 \pm 2.09$} & - & $13.16 \pm 3.83$\\
  rc\_207.3 & 33 & \bm{$9.63 \pm 0.81$} & $12.52 \pm 3.18$ & - & $9.94 \pm 0.78$\\
  rc\_207.2 & 31 & - & $11.74 \pm 2.48$ & - & \bm{$8.65 \pm 2.18$}\\
  rc\_206.3 & 25 & $9.07 \pm 1.83$ & $9.63 \pm 2.02$ & - & \bm{$7.58 \pm 1.84$}\\
  rc\_207.1 & 34 & $7.73 \pm 1.11$ & $11.28 \pm 1.72$ & - & \bm{$7.68 \pm 0.93$}\\
  rc\_202.1 & 33 & - & $8.73 \pm 6.31$ & - & \bm{$8.12 \pm 1.06$} \\
  rc\_205.3 & 35 & $13.46 \pm 4.78$ & $13.09 \pm 2.69$ & - & \bm{$11.99 \pm 2.43$}\\
  rc\_201.1 & 20 & \bm{$4.44 \pm 0.01$} & \bm{$4.44 \pm 0.01$} & $13.65 \pm 2.17$ & \bm{$4.57 \pm 0.08$}\\
  rc\_205.2 & 27 & - & - & - & \bm{$9.07 \pm 0.69$} \\
  rc\_205.4 & 28 & $9.50 \pm 4.52$ & $8.49 \pm 2.26$ & - & \bm{$7.11 \pm 0.76$}\\
  rc\_202.3 & 29 & - & - & - & \bm{$6.51 \pm 0.63$}\\
  rc\_206.2 & 37 & - & - & - & \bm{$12.56 \pm 2.56$} \\
  rc\_206.4 & 38 & - & - & - & \bm{$7.20 \pm 2.16$}\\
  rc\_201.2 & 26 &  $6.79 \pm 0.30$&  $8.54 \pm 2.17$& - & \bm{$6.57 \pm 0.44$}\\
  rc\_201.4 & 26 & - & \bm{$13.48 \pm 11.13$} & - & $19.73 \pm 0.10$\\
  rc\_201.3 & 32 & - & - & - & \bm{$4.38 \pm 0.48$}\\

  \bottomrule
\end{tabular}
\end{table}

\begin{table}
\centering
\renewcommand{\arraystretch}{1.6}
\caption{Average constraint violations after 100000 objective function evaluations on all instances of MO-TSPTW}
\label{table:full-tsptw-success}
\begin{tabular}{lccccr}
  \toprule
  Instance & Cities & NSGA-II & SMS-EMOA & Pareto-MCTS & Pareto-NRPA \\
  \midrule
  rc\_206.1 &  4 & \bm{$0.00 \pm 0.00$} & \bm{$0.00 \pm 0.00$} & \bm{$0.00 \pm 0.00$} & \bm{$0.00 \pm 0.00$} \\
  rc\_207.4 &  6 & \bm{$0.00 \pm 0.00$} & \bm{$0.00 \pm 0.00$} & \bm{$0.00 \pm 0.00$} & \bm{$0.00 \pm 0.00$} \\
  rc\_203.4 & 15 & \bm{$0.00 \pm 0.00$} & \bm{$0.00 \pm 0.00$} & $0.00 \pm 0.00$ & \bm{$0.00 \pm 0.00$} \\
  rc\_202.2 & 14 & \bm{$0.00 \pm 0.00$} & \bm{$0.00 \pm 0.00$} & \bm{$0.00 \pm 0.00$} & \bm{$0.00 \pm 0.00$} \\
  rc\_204.3 & 24 & \bm{$0.00 \pm 0.00$} & \bm{$0.00 \pm  0.00$} & $0.00 \pm 0.00$ & \bm{$0.00 \pm 0.00$} \\
  rc\_203.1 & 19 & \bm{$0.00 \pm 0.00$} & \bm{$0.00 \pm 0.00$} & \bm{$0.00 \pm 0.00$} & \bm{$0.00 \pm 0.00$} \\
  rc\_204.2 & 33 & \bm{$0.00 \pm 0.00$} & \bm{$0.00 \pm 0.00$} & $4.07 \pm 0.28$ & \bm{$0.00 \pm 0.00$} \\
  rc\_208.2 & 29 & \bm{$0.00 \pm 0.00$} & \bm{$0.00 \pm 0.00$} & $7.10 \pm 0.41$ & \bm{$0.00 \pm 0.00$} \\
  rc\_204.4 & 14 & \bm{$0.00 \pm 0.00$} & \bm{$0.00 \pm 0.00$} & \bm{$0.00 \pm 0.00$} & \bm{$0.00 \pm 0.00$} \\
  rc\_205.1 & 14 & \bm{$0.00 \pm 0.00$} & \bm{$0.00 \pm 0.00$} & \bm{$0.00 \pm 0.00$} & \bm{$0.00 \pm 0.00$} \\
  rc\_204.1 & 46 & $1.90 \pm 0.47$ & $2.17 \pm 0.43$ & $20.47 \pm 0.32$ & \bm{$0.00 \pm 0.00$}\\
  rc\_203.2 & 33 & $1.90 \pm 0.54$ & $1.17 \pm 0.43$ & $10.63 \pm 0.45$ & \bm{$0.00 \pm 0.00$}\\
  rc\_203.3 & 37 & $2.40 \pm 0.52$ & $2.63 \pm 0.43$ & $10.53 \pm 0.26$ & \bm{$0.00 \pm 0.00$} \\
  rc\_208.3 & 36 & $0.03 \pm 0.06$ & \bm{$0.00 \pm 0.00$} & $13.43 \pm 0.42$ & \bm{$0.00 \pm 0.00$} \\
  rc\_202.4 & 28 & $1.57 \pm 0.30$ & $2.10 \pm 0.55$ & $6.60 \pm 0.35$ & \bm{$0.07 \pm 0.09$} \\
  rc\_208.1 & 38 & $0.97 \pm 0.15$ & $1.03 \pm 0.22$ & $23.97 \pm 0.22$ & \bm{$0.13 \pm 0.26$}\\
  rc\_207.3 & 33 & $1.23 \pm 0.41$ & $1.13 \pm 0.35$ & $16.03 \pm 0.35$ & \bm{$0.00 \pm 0.00$}\\
  rc\_207.2 & 31 & $1.17 \pm 0.16$ & $1.03 \pm 0.17$ & $18.50 \pm 0.20$ & \bm{$0.40 \pm 0.18$} \\
  rc\_206.3 & 25 & $0.67 \pm 0.35$ & $0.50 \pm 0.33$ & $1.00 \pm 0.13$ & \bm{$0.00 \pm 0.00$}\\
  rc\_207.1 & 34 & $0.73 \pm 0.33$ & $0.70 \pm 0.34$ & $4.60 \pm 0.46$ & \bm{$0.00 \pm 0.00$}\\
  rc\_202.1 & 33 & $1.93 \pm 0.32$ & $1.70 \pm 0.44$ & $9.00 \pm 0.49$ & \bm{$0.00 \pm 0.00$} \\
  rc\_205.3 & 35 & $1.90 \pm 0.55$ & $1.80 \pm 0.53$ & $10.80 \pm 0.33$ & \bm{$0.00 \pm 0.00$}\\
  rc\_201.1 & 20 & $0.20 \pm 0.14$ & $0.33 \pm 0.21$ & \bm{$0.00 \pm 0.00$} & \bm{$0.00 \pm 0.00$} \\
  rc\_205.2 & 27 & $4.73 \pm 0.51$ & $3.90 \pm 0.44$ & $3.83 \pm 0.31$ & \bm{$0.00 \pm 0.00$}\\
  rc\_205.4 & 28 & $0.93 \pm 0.32$ & $0.87 \pm 0.35$ & $5.13 \pm 0.29$ & \bm{$0.00 \pm 0.00$} \\
  rc\_202.3 & 29 & $2.10 \pm 0.47$ & $2.63 \pm 0.60$ & $10.23 \pm 0.27$ & \bm{$0.00 \pm 0.00$} \\
  rc\_206.2 & 37 & $4.70 \pm 0.63$ & $3.70 \pm 0.55$ & $22.17 \pm 0.29$ & \bm{$0.03 \pm 0.06$}\\
  rc\_206.4 & 38 & $5.13 \pm 0.47$ & $4.57 \pm 0.49$ & $19.13 \pm 0.27$ & \bm{$0.10 \pm 0.14$}\\
  rc\_201.2 & 26 & $3.63 \pm 0.54$ & $1.93 \pm 0.73$ & $4.37 \pm 0.24$ & \bm{$0.00 \pm 0.00$}\\
  rc\_201.4 & 26 & $2.97 \pm 0.54$ & $1.93 \pm 0.36$ & $3.40 \pm 0.25$ & \bm{$0.00 \pm 0.00$} \\
  rc\_201.3 & 32 & $7.33 \pm 0.50$ & $5.97 \pm 0.43$ & $9.63 \pm 0.52$ & \bm{$0.00 \pm 0.00$}\\
  
  \bottomrule
\end{tabular}

\end{table}

\FloatBarrier

\subsection{Impact of bias on evolutionary multi-objective optimization algorithms} \label{sec:ablation-emoa-bias}

\begin{figure}
  \centering
  \subfigure[Aggregated Pareto front]{\includegraphics[width=.48\linewidth]{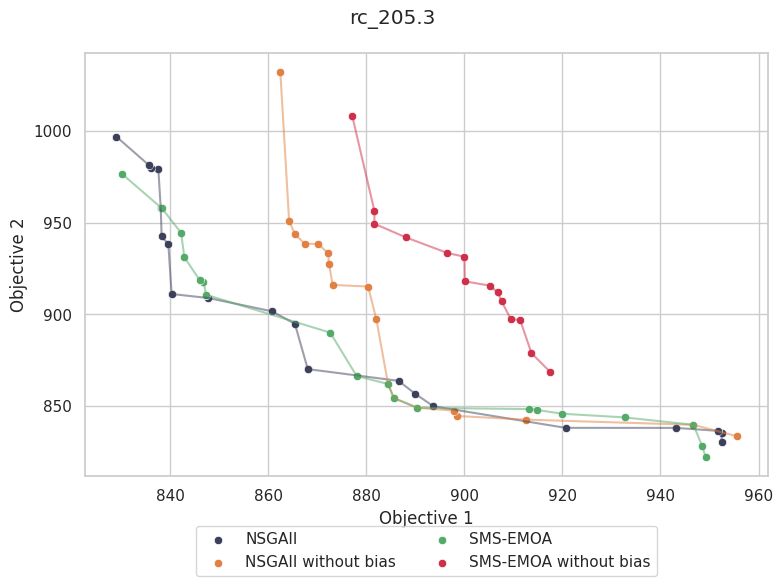}} \hfill
  \subfigure[Normalized hypervolume evolution]{\includegraphics[width=.48\linewidth]{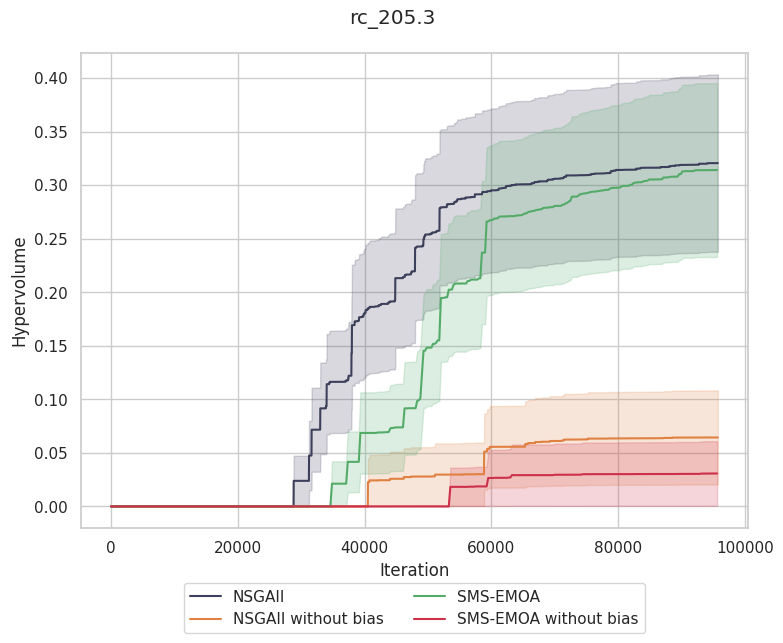}}
  \caption{Impact of bias on EMOA convergence on rc\_205.3}
  \label{fig:bias-nsga2}
\end{figure}

Section \ref{sec:results} indicates that implementing a bias term in the rollout policy strongly improves NRPA performance. In order to provide a fair comparison between Pareto-NRPA and state-of-the-art MOO algorithms, the same bias term is implemented in the sampling operators for NSGA-II and SMS-EMOA. Figures \ref{fig:bias-nsga2}a and \ref{fig:bias-nsga2}b as well as Table \ref{table:hv-bias-emoa} show that implementing a bias term improves the results of NSGA-II as well. The initialization part of the algorithm benefits from higher-quality samples thanks to the bias, which gives an important head start to algorithm convergence. Results in Table \ref{table:hv-bias-emoa} demonstrate that adding a bias to NSGA-II gives an equal or higher hypervolume in 23 out of 31 instances, and adding a bias to SMS-EMOA gives equal or higher hypervolume in 25 out of 31 instances. 

\begin{table}
\centering
\renewcommand{\arraystretch}{1.6}
\caption{Impact of bias on EMOA hypervolume}
\label{table:hv-bias-emoa}
\begin{tabular}{lccccr}
  \toprule
  Instance & Cities & NSGA-II & NSGA-II (no bias) & SMS-EMOA & SMS-EMOA (no bias) \\
  \midrule
  rc\_206.1 &  4 & \bm{$1.00 \pm 0.00$} & \bm{$1.00 \pm 0.00$} & \bm{$1.00 \pm 0.00$} & \bm{$1.00 \pm 0.00$}\\
  rc\_207.4 &  6 & \bm{$1.00 \pm 0.00$} & \bm{$1.00 \pm 0.00$} & \bm{$1.00 \pm 0.00$} & \bm{$1.00 \pm 0.00$}\\
  rc\_203.4 & 15 & \bm{$1.00 \pm 0.00$} & $0.97 \pm 0.01$ &\bm{$1.00 \pm 0.00$} & $0.96 \pm 0.02$ \\
  rc\_202.2 & 14 & \bm{$1.00 \pm 0.00$} & $0.97 \pm 0.02$ &\bm{$1.00 \pm 0.00$} & $0.98 \pm 0.00$ \\
  rc\_204.3 & 24 & \bm{$0.97 \pm 0.01$} & $0.96 \pm 0.01$ &$0.96 \pm 0.01$ &$0.95 \pm 0.01$ \\
  rc\_203.1 & 19 & \bm{$0.91 \pm 0.02$} & $0.78 \pm 0.03$ &$0.87 \pm 0.03$ &$0.77 \pm 0.02$ \\
  rc\_204.2 & 33 & \bm{$0.80 \pm 0.03$} &$0.70 \pm 0.06$ &\bm{$0.80 \pm 0.04$} & $0.68 \pm 0.05$ \\
  rc\_208.2 & 29 & \bm{$0.94 \pm 0.01$} & \bm{$0.94 \pm 0.01$} & $0.93 \pm 0.01$ &$0.92 \pm 0.01$ \\
  rc\_204.4 & 14 & \bm{$1.00 \pm 0.00$} & \bm{$1.00 \pm 0.00$} & \bm{$1.00 \pm 0.00$} & \bm{$1.00 \pm 0.00$}\\
  rc\_205.1 & 14 & \bm{$1.00 \pm 0.00$} & $0.99 \pm 0.01$ &\bm{$1.00 \pm 0.00$} & \bm{$1.00 \pm 0.00$} \\
  rc\_204.1 & 46 & \bm{$0.12 \pm 0.11$} & $0.00 \pm 0.00$ &$0.00 \pm 0.00$ &$0.00 \pm 0.00$ \\
  rc\_203.2 & 33 & $0.11 \pm 0.08$ &$0.24 \pm 0.12$ &\bm{$0.27 \pm 0.12$} & $0.19 \pm 0.11$ \\
  rc\_203.3 & 37 & \bm{$0.09 \pm 0.07$} & $0.01 \pm 0.02$ &$0.05 \pm 0.05$ &$0.00 \pm 0.00$ \\
  rc\_208.3 & 36 & $0.85 \pm 0.06$ &$0.79 \pm 0.06$ &\bm{$0.86 \pm 0.02$} & $0.80 \pm 0.03$ \\
  rc\_202.4 & 28 & $0.02 \pm 0.02$ &$0.10 \pm 0.07$ &$0.07 \pm 0.07$ &\bm{$0.18 \pm 0.10$} \\
  rc\_208.1 & 38 & $0.06 \pm 0.07$ &\bm{$0.10 \pm 0.09$} & $0.06 \pm 0.07$ &$0.10 \pm 0.08$ \\
  rc\_207.3 & 33 & $0.42 \pm 0.15$ &$0.29 \pm 0.14$ &\bm{$0.44 \pm 0.14$} & \bm{$0.44 \pm 0.14$} \\
  rc\_207.2 & 31 & $0.00 \pm 0.00$ &$0.11 \pm 0.10$ &$0.08 \pm 0.09$ &\bm{$0.13 \pm 0.10$} \\
  rc\_206.3 & 25 & $0.56 \pm 0.17$ &$0.70 \pm 0.15$ &$0.63 \pm 0.16$ &\bm{$0.81 \pm 0.11$} \\
  rc\_207.1 & 34 & $0.42 \pm 0.15$ &$0.29 \pm 0.14$ &\bm{$0.44 \pm 0.14$} & \bm{$0.44 \pm 0.14$} \\
  rc\_202.1 & 33 & $0.00 \pm 0.00$ &$0.03 \pm 0.05$ &\bm{$0.08 \pm 0.09$} & $0.00 \pm 0.00$ \\
  rc\_205.3 & 35 & \bm{$0.27 \pm 0.14$} & $0.05 \pm 0.07$ &$0.25 \pm 0.13$ &$0.02 \pm 0.04$ \\
  rc\_201.1 & 20 & \bm{$0.77 \pm 0.15$} & $0.60 \pm 0.18$ &$0.73 \pm 0.16$ &$0.60 \pm 0.18$ \\
  rc\_205.2 & 27 & $0.00 \pm 0.00$ &$0.03 \pm 0.06$ &$0.00 \pm 0.00$ &\bm{$0.09 \pm 0.09$} \\
  rc\_205.4 & 28 & $0.26 \pm 0.13$ &$0.16 \pm 0.11$ &\bm{$0.28 \pm 0.14$} & $0.16 \pm 0.11$ \\
  rc\_202.3 & 29 & $0.00 \pm 0.00$ &$0.00 \pm 0.00$ &$0.00 \pm 0.00$ &\bm{$0.03 \pm 0.06$} \\
  rc\_206.2 & 37 & $0.00 \pm 0.00$ & $0.00 \pm 0.00$ & $0.00 \pm 0.00$ & $0.00 \pm 0.00$ \\
  rc\_206.4 & 38 & $0.00 \pm 0.00$ &\bm{$0.02 \pm 0.04$} &$0.00 \pm 0.00$ &$0.00 \pm 0.00$ \\
  rc\_201.2 & 26 & $0.06 \pm 0.08$ &$0.09 \pm 0.10$ &\bm{$0.25 \pm 0.15$} &$0.03 \pm 0.06$ \\
  rc\_201.4 & 26 & $0.00 \pm 0.00$ &$0.00 \pm 0.00$ &\bm{$0.07 \pm 0.09$} &\bm{$0.07 \pm 0.09$} \\
  rc\_201.3 & 32 & $0.00 \pm 0.00$ & $0.00 \pm 0.00$ & $0.00 \pm 0.00$ & $0.00 \pm 0.00$ \\
  
  \bottomrule
\end{tabular}

\end{table}

\FloatBarrier
\newpage

\subsection{Ablation study: Adapt algorithm} \label{sec:ablation-adapt}

We are interested in finding out if adapting the policy with respect to all sequences in the Pareto front performs better than with only one sequence. In the \textit{one sequence} setting, the solutions of the Pareto front are sorted according to their crowding distance, and each policy $\pi_k$ updates with respect to the sequence originating from $\pi_k$ that maximizes the crowding distance. The two algorithms are run on the instance \texttt{rc\_204.3}. An easy instance such as this one is chosen because NRPA quickly finds multiple valid solutions, which highlights the difference between the two adapt methods.

\begin{table}
\centering
\renewcommand{\arraystretch}{1.4}
\caption{Metrics for the two adapt strategies}
\label{table:adapt-one-all}
\begin{tabular}{lcccr}
\toprule
Adapt strategy & Normalized hypervolume & Overall Spread & Spacing &  \\
\midrule
One sequence   & $0.50 \pm 0.10$        & $0.25 \pm 0.04$       & $10.48 \pm 2.16$ \\
All sequences  & \bm{$0.70 \pm 0.09$}  & \bm{$0.56 \pm 0.07$} & \bm{$7.63 \pm 1.20$} \\
\bottomrule
\end{tabular}
\end{table}

\begin{figure}
  \centering
  \subfigure[Aggregated Pareto front for the two adapt strategies]{\includegraphics[width=.48\linewidth]{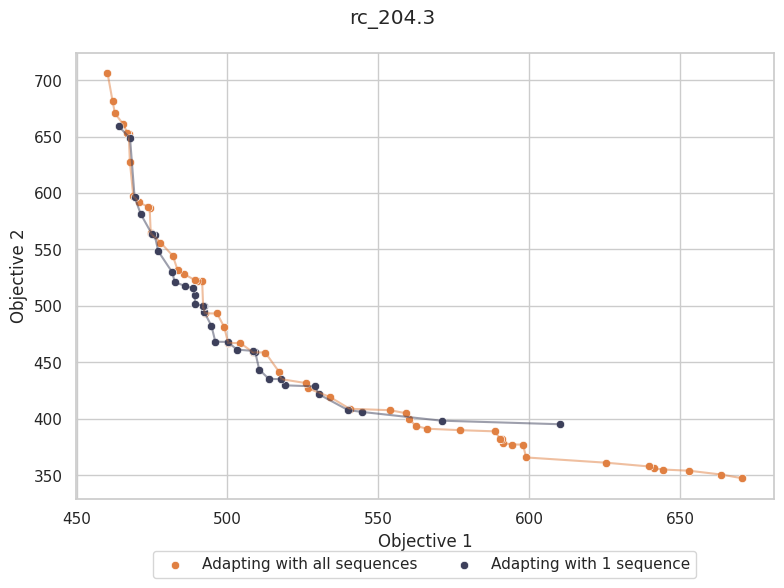}} \hfill
  \subfigure[Normalized hypervolume evolution for the two adapt strategies]{\includegraphics[width=.48\linewidth]{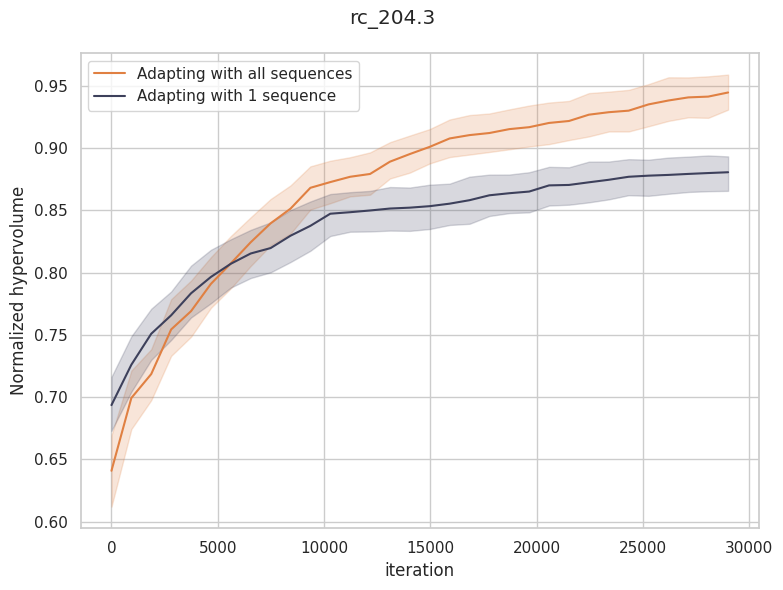}}
  \caption{One-vs-all policy adaptation method}
  \label{fig:adapt-one-all}
\end{figure}

Figures \ref{fig:adapt-one-all}a, \ref{fig:adapt-one-all}b and Table \ref{table:adapt-one-all} show the differences between adapting the policy with respect to a single sequence compared to all sequences. The metrics are shown with a 95\% confidence interval. It is clear that adapting with all sequences gives the algorithm a significant performance boost, not only in solution quality but also diversity.

\FloatBarrier
\newpage

\subsection{Ablation study: Crowding distance weighting} \label{sec:ablation-crowding}

Section \ref{sec:contributions} mentions the use of crowding distance weighting during policy adaptation. Indeed, Algorithm \ref{alg:pareto-adapt} shows that the gradient ascent step for policy $\pi_k$ and sequence $s \in S^* | s.policy = \pi_k $ is weighted by a quantity related to the crowding distance (CD) \cite{deb_fast_2002} of the sequence compared to other individuals in $S^*$. This design choice is made to promote policy adaptation towards more isolated sequences, with the aim to maximize the coverage of the Pareto front approximation in the solution space. In this ablation study, we compare Pareto-NRPA as presented in Section \ref{sec:contributions} to Pareto-NRPA where each sequence has the same weight during policy adaptation (Pareto-NRPA without CD weighting). The relevant metric for this comparison is the overall spread (OS) metric, which indicates the extent of the Pareto front approximation. Table \ref{table:crowding-nrpa} clearly shows that using crowding distance weighting during the Pareto-Adapt algorithm significantly improves the spread of $S^*$ in the solution space.
\begin{table}[]
\centering
\renewcommand{\arraystretch}{1.6}
\caption{Overall spread for Pareto-NRPA with and without CD weighting}
\label{table:crowding-nrpa}

\begin{tabular}{lccccr}
  \toprule
  Instance & Cities & Pareto-NRPA & Pareto-NRPA without CD weighting \\
  \midrule
  rc\_206.1 & 4  & \bm{$1.00 \pm 0.00$} & \bm{$1.00 \pm 0.00$} \\
  rc\_207.4 &  6 & \bm{$1.00 \pm 0.00$} & \bm{$1.00 \pm 0.00$} \\
  rc\_203.4 & 15 & $0.51 \pm 0.05$ & \bm{$0.56 \pm 0.04$} \\
  rc\_202.2 & 14 & $0.53 \pm 0.00$ & \bm{$0.56 \pm 0.04$} \\
  rc\_204.3 & 24 & \bm{$0.67 \pm 0.03$} & $0.53 \pm 0.07$ \\
  rc\_203.1 & 19 & \bm{$0.65 \pm 0.06$} & $0.58 \pm 0.08$ \\
  rc\_204.2 & 33 & \bm{$0.47 \pm 0.04$} & $0.28 \pm 0.04$ \\
  rc\_208.2 & 29 & \bm{$0.38 \pm 0.05$} & $0.21 \pm 0.03$ \\
  rc\_204.4 & 14 & $0.88 \pm 0.02$ & \bm{$0.90 \pm 0.01$} \\
  rc\_205.1 & 14 & \bm{$0.93 \pm 0.04$} & $0.90 \pm 0.04$ \\
  rc\_204.1 & 46 & \bm{$0.17 \pm 0.06$} & $0.00 \pm 0.00$ \\
  rc\_203.2 & 33 & \bm{$0.50 \pm 0.04$} & $0.38 \pm 0.04$ \\
  rc\_203.3 & 37 & \bm{$0.38 \pm 0.05$} & $0.22 \pm 0.04$ \\
  rc\_208.3 & 36 & \bm{$0.26 \pm 0.04$} & $0.09 \pm 0.02$ \\
  rc\_202.4 & 28 & \bm{$0.40 \pm 0.06$} & $0.35 \pm 0.06$ \\
  rc\_208.1 & 38 & \bm{$0.15 \pm 0.04$} & $0.04 \pm 0.02$ \\
  rc\_207.3 & 33 & \bm{$0.58 \pm 0.06$} & $0.37 \pm 0.04$ \\
  rc\_207.2 & 31 & \bm{$0.10 \pm 0.04$} & $0.04 \pm 0.02$ \\
  rc\_206.3 & 25 & \bm{$0.53 \pm 0.07$} & $0.42 \pm 0.05$ \\
  rc\_207.1 & 34 & \bm{$0.43 \pm 0.05$} & $0.29 \pm 0.06$ \\
  rc\_202.1 & 33 & \bm{$0.44 \pm 0.06$} & $0.21 \pm 0.04$ \\
  rc\_205.3 & 35 & \bm{$0.19 \pm 0.04$} & $0.17 \pm 0.04$ \\
  rc\_201.1 & 20 & \bm{$0.57 \pm 0.08$} & $0.42 \pm 0.07$ \\
  rc\_205.2 & 27 & \bm{$0.65 \pm 0.06$} & $0.58 \pm 0.05$ \\
  rc\_205.4 & 28 & $0.49 \pm 0.05$ & \bm{$0.55 \pm 0.05$} \\
  rc\_202.3 & 29 & \bm{$0.71 \pm 0.06$} & $0.65 \pm 0.06$ \\
  rc\_206.2 & 37 & \bm{$0.21 \pm 0.05$} & $0.16 \pm 0.04$ \\
  rc\_206.4 & 38 & \bm{$0.11 \pm 0.03$} & $0.03 \pm 0.01$ \\
  rc\_201.2 & 26 & \bm{$0.49 \pm 0.09$} & $0.47 \pm 0.08$ \\
  rc\_201.4 & 26 & \bm{$0.48 \pm 0.02$} & $0.44 \pm 0.05$ \\
  rc\_201.3 & 32 & \bm{$0.34 \pm 0.03$} & $0.26 \pm 0.03$ \\
  
  \bottomrule
\end{tabular}

\end{table}

\FloatBarrier
\newpage

\subsection{Hyperparameter sensitivity analysis} \label{sec:hyperparameter}

In order to choose the hyperparameters for Pareto-NRPA, we run a concise grid search on the rc\_205.2 instance. The goal is not to find the best hyperparameters for each instance, but rather to identify a suitable hyperparameter configuration for all instances. As such, finding the optimal parameter combination for this particular instance is not the aim of this grid search. The hyperparameter search is conducted over values of $\alpha = \{0.1, 0.25, 0.5, 0.75, 1, 2\}$ and values of $|\Pi| = \{1, 2, 4\}$. Once hyperparameter values have been identified ($\alpha = 0.5$ and $|\Pi| = 4$), the $level$ parameter is studied for values in $\{3, 4 \}$.

\begin{table}
\centering
\renewcommand{\arraystretch}{1.1}
\caption{Normalized hypervolume for each hyperparameter configuration.}
\label{table:hv-grid}
\begin{tabular}{lcccccc}
\toprule
$\Pi \backslash \alpha$ & 0.1 & 0.25 & 0.5 & 0.75 & 1 & 2 \\
\midrule
1 & $0.92 \pm 0.04$ & $0.83 \pm 0.11$ & $0.67 \pm 0.13$ & $0.46 \pm 0.13$ & $0.50 \pm 0.14$ & $0.38 \pm 0.13$ \\
2 & $0.88 \pm 0.04$ & $0.92 \pm 0.04$ & $0.92 \pm 0.04$ & $0.87 \pm 0.07$ & $0.66 \pm 0.13$ & $0.40 \pm 0.13$ \\
4 & $0.84 \pm 0.02$ & \bm{$0.93 \pm 0.02$} & \bm{$0.93 \pm 0.02$} & $0.82 \pm 0.10$ & $0.83 \pm 0.07$ & $0.65 \pm 0.10$ \\
\bottomrule
\end{tabular}

\caption{Overall spread for each hyperparameter configuration.}
\label{table:os-grid}
\begin{tabular}{lcccccc}
\toprule
$\Pi \backslash \alpha$ & 0.1 & 0.25 & 0.5 & 0.75 & 1 & 2 \\
\midrule
1 & $0.49 \pm 0.05$ & $0.42 \pm 0.07$ & $0.32 \pm 0.08$ & $0.20 \pm 0.08$ & $0.23 \pm 0.08$ & $0.14 \pm 0.06$ \\
2 & $0.40 \pm 0.05$ & $0.53 \pm 0.05$ & $0.53 \pm 0.05$ & $0.43 \pm 0.06$ & $0.31 \pm 0.08$ & $0.16 \pm 0.06$ \\
4 & $0.28 \pm 0.05$ & \bm{$0.54 \pm 0.05$} & \bm{$0.54 \pm 0.05$} & $0.45 \pm 0.08$ & $0.40 \pm 0.05$ & $0.28 \pm 0.06$ \\
\bottomrule
\end{tabular}

\caption{Spacing for each hyperparameter configuration.}
\label{table:spacing-grid}
\begin{tabular}{lcccccc}
\toprule
$\Pi \backslash \alpha$ & 0.1 & 0.25 & 0.5 & 0.75 & 1 & 2 \\
\midrule
1 & $8.65 \pm 0.58$ & $7.64 \pm 0.69$ & $6.66 \pm 0.93$ & $6.17 \pm 1.26$ & $7.62 \pm 1.53$ & \bm{$5.54 \pm 1.53$} \\
2 & $8.40 \pm 0.73$ & $8.42 \pm 0.60$ & $8.09 \pm 0.60$ & $8.09 \pm 0.88$ & $7.70 \pm 1.97$ & $6.81 \pm 1.44$ \\
4 & $8.71 \pm 0.68$ & $8.94 \pm 0.57$ & $8.85 \pm 0.56$ & $8.24 \pm 0.64$ & $8.18 \pm 0.85$ & $8.84 \pm 1.60$ \\
\bottomrule
\end{tabular}

\caption{Average constraint violations for each hyperparameter configuration.}
\label{table:cv-grid}
\begin{tabular}{lcccccc}
\toprule
$\Pi \backslash \alpha$ & 0.1 & 0.25 & 0.5 & 0.75 & 1 & 2 \\
\midrule
1 & \bm{$0.00 \pm 0.00$} & $0.03 \pm 0.06$ & $0.07 \pm 0.09$ & $0.20 \pm 0.08$ & $0.23 \pm 0.08$ & $0.14 \pm 0.06$ \\
2 & \bm{$0.00 \pm 0.00$} & \bm{$0.00 \pm 0.00$} & \bm{$0.00 \pm 0.00$} & \bm{$0.00 \pm 0.00$} & $0.13 \pm 0.12$ & $0.43 \pm 0.20$ \\
4 & \bm{$0.00 \pm 0.00$} & \bm{$0.00 \pm 0.00$} & \bm{$0.00 \pm 0.00$} & $0.04 \pm 0.08$ & $0.03 \pm 0.06$ & $0.04 \pm 0.08$ \\
\bottomrule
\end{tabular}

\caption{Metrics on rc\_205.2 for different values of NRPA level}
\label{table:level-nrpa}
\begin{tabular}{lcccr}
\toprule
Algorithm & Hypervolume & Overall Spread & Spacing & CV \\
\midrule
Pareto-NRPA level 3 & $0.86 \pm 0.05$ & $0.48 \pm 0.06$ & \bm{$8.82 \pm 0.79$} & \bm{$0.00 \pm 0.00$} \\
Pareto-NRPA level 4 & \bm{$0.94 \pm 0.03$} & \bm{$0.63 \pm 0.05$} & $8.95 \pm 0.57$ & \bm{$0.00 \pm 0.00$} \\
\bottomrule
\end{tabular}
\end{table}

\begin{figure}
    \centering
    \subfigure[]{\includegraphics[height=4cm]{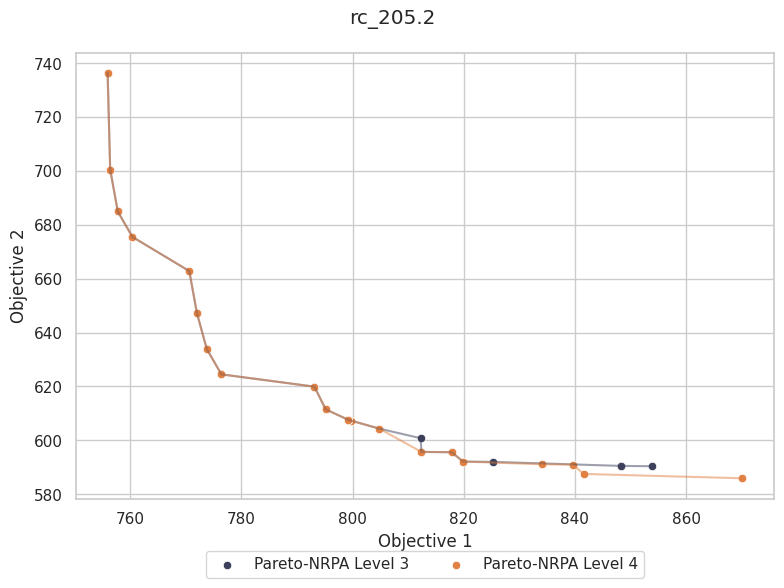}} 
    \subfigure[]{\includegraphics[height=4cm]{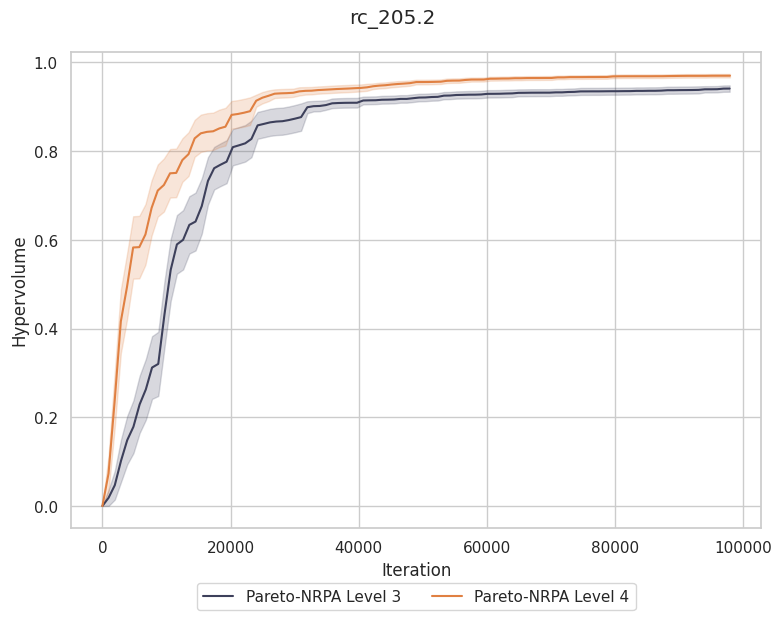}}
    \caption{Aggregated Pareto front and hypervolume evolution for different values of $level$}
    \label{fig:enter-label}
\end{figure}

\subsection{Results on the unconstrained multi-objective traveling salesman problem}

This paper introduces a novel benchmark dataset, MO-TSPTW, based on classical instances of the traveling salesman problem with time windows. As such, the main experiments in the paper reflect performance on this new dataset. However, one might argue that benchmarking Pareto-NRPA on the better-known MO-TSP problem is of interest. MO-TSP associates two independent costs matrices to a number $n$ of cities, which can be visited in any order as there aren't any time windows to respect.

We benchmark Pareto-NRPA on two MO-TSP problem sizes: 50 cities and 100 cities. It may be appreciated that both of these sizes are larger than the largest MO-TSPTW instance (46 cities) but remain relatively smal. Indeed, Pareto-NRPA becomes computationally inefficient when the number of cities is larger than 500, due to the policy vector exploding in size. Improving the computational efficiency of Pareto-NRPA remains an avenue for future research.

\begin{table}[H]
\centering
\renewcommand{\arraystretch}{1.4}
\caption{MO-TSP : 50 cities}
\label{table:motsp-50}
\begin{tabular}{lccr}
\toprule
Algorithm & Hypervolume & Overall Spread & Spacing \\
\midrule
NSGA-II     & $0.88 \pm 0.00$      & $0.51 \pm 0.03$      & $682.08 \pm 87.39$  \\
SMS-EMOA    & $0.85 \pm 0.00$      & $0.35 \pm 0.03$      & $828.07 \pm 115.47$       \\
PLS         & $0.88 \pm 0.01$      & $0.22 \pm 0.02$      & \bm{$429.07 \pm 34.41$}       \\
MOEA/D      & $0.93 \pm 0.00$      & \bm{$0.55 \pm 0.03$}      & $589.00 \pm 68.94$      \\
Pareto-NRPA & \bm{$0.98 \pm 0.00$} & $0.41 \pm 0.02$ & $828.07 \pm 115.47$      \\ 
\bottomrule
\end{tabular}
\end{table}

\begin{table}[H]
\centering
\renewcommand{\arraystretch}{1.4}
\caption{MO-TSP : 100 cities}
\label{table:motsp-100}
\begin{tabular}{lccr}
\toprule
Algorithm & Hypervolume & Overall Spread & Spacing \\
\midrule
NSGA-II     & $0.79 \pm 0.01$      & $0.25 \pm 0.10$      & \bm{$711.67 \pm 86.60$}  \\
SMS-EMOA    & $0.77 \pm 0.01$      & $0.34 \pm 0.05$      & $853.00 \pm 94.88$       \\
PLS         & $0.80 \pm 0.01$      & $0.16 \pm 0.02$      & $561.86 \pm 56.71$       \\
MOEA/D      & $0.85 \pm 0.01$      & $0.32 \pm 0.11$      & $783.72 \pm 136.09$      \\
Pareto-NRPA & \bm{$0.97 \pm 0.01$} & \bm{$0.42 \pm 0.04$} & $849.45 \pm 114.69$      \\ 
\bottomrule
\end{tabular}
\end{table}

On the two MO-TSP datasets, Pareto-NRPA yields the best hypervolume values after 30 runs of 100000 iterations. We note that the performance of Pareto-NRPA is much closer to the other state-of-the-art algorithms on MO-TSP. This shows that MO-TSPTW exhibits higher complexity due to the time window constraints. As such, we believe that MO-TSPTW may be used for future research.

\begin{figure}
  \centering
  \subfigure[MO-TSP with 50 cities]{\includegraphics[width=.48\linewidth]{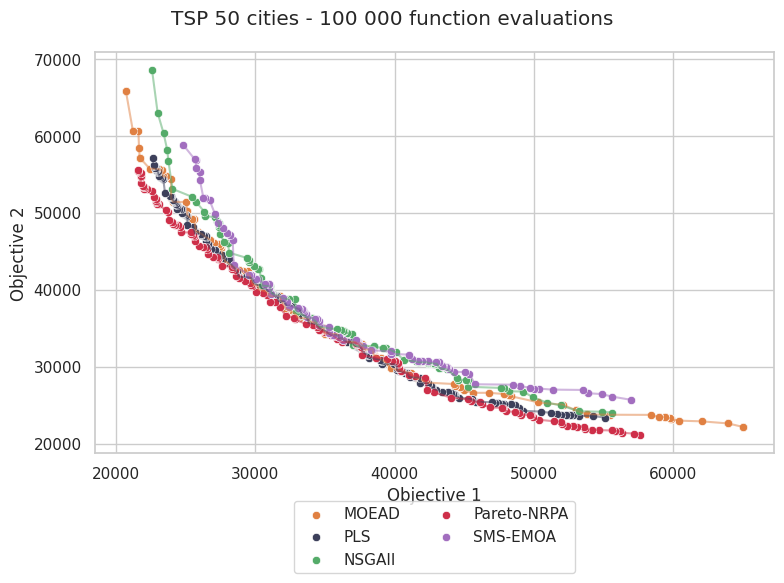}} \hfill
  \subfigure[MO-TSP with 100 cities]{\includegraphics[width=.48\linewidth]{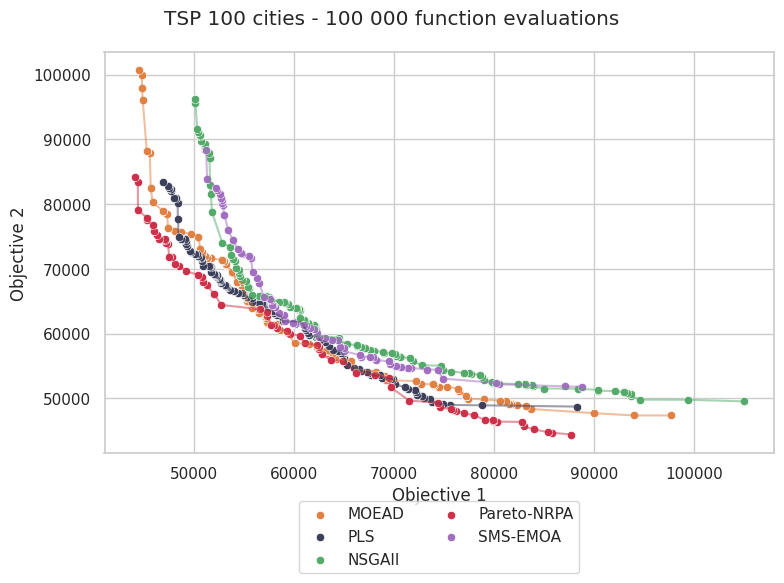}}
  \caption{Pareto front approximations on MO-TSP}
  \label{fig:adapt-one-all}
\end{figure}

\newpage

\subsection{On algorithm compexity and CPU time}

Despite its remarkable sample effiency and fast convergence speed, it may be observed that Pareto-NRPA's algorithmic complexity increases with the size of the search space. Indeed, the Adapt algorithm requires copying the policy vector to a temporary value. When the search space is large, the number of stat-action couples greatly increases and, as such, so does the cost of copying the policy. For example, copying the policy vector for MO-TSP with 500 cities exhibits a redhibitory computational cost. 

Given the fact that Pareto-NRPA is less time-efficient than competing state-of-the-art algorithms, we benchmark the performances under the constraint of equal CPU-time for all algorithms, instead of equal number of function evaluations. This scenario is more representative of problems where evaluating the objectives is quick, while fixing the number of function evaluations represents cases (such as neural architecture search) where evaluating a solution is costly.

\begin{table}[H]
\footnotesize
\centering
\renewcommand{\arraystretch}{1.4}
\caption{Metrics on rc\_204.3}
\label{table:metrics-rc204-3-equal-cpu}
\begin{tabular}{lcccr}
\toprule
Algorithm & Hypervolume & Overall Spread & Spacing & CV\\
\midrule
NSGA-II     & $0.97 \pm 0.02$      & $0.43 \pm 0.04$      & $6.13 \pm 2.25$      & \bm{$0.00 \pm 0.00$} \\
SMS-EMOA    & $0.75 \pm 0.35$      & $0.27 \pm 0.15$      & $11.08 \pm 12.35$    & \bm{$0.00 \pm 0.00$} \\
Pareto-MCTS & $0.41 \pm 0.03$      & $0.23 \pm 0.08$      & $44.94 \pm 24.52$    & \bm{$0.00 \pm 0.00$} \\
PLS         & \bm{$0.99 \pm 0.01$} & \bm{$0.68 \pm 0.18$} & \bm{$3.69 \pm 1.41$} & \bm{$0.00 \pm 0.00$} \\
MOEA/D      & $0.96 \pm 0.03$      & $0.35 \pm 0.07$      & $4.57 \pm 2.06$      & \bm{$0.00 \pm 0.00$} \\
Pareto-NRPA & $0.65 \pm 0.15$      & $0.15 \pm 0.07$      & $8.82 \pm 2.11$      & \bm{$0.00 \pm 0.00$} \\
\bottomrule
\end{tabular}
\end{table}

\begin{table}[H]
\centering
\renewcommand{\arraystretch}{1.4}
\caption{Metrics on rc\_201.3}
\label{table:metrics-rc201-3-equal-cpu}
\begin{tabular}{lcccr}
\toprule
Algorithm & Hypervolume & Overall Spread & Spacing & CV \\
\midrule
NSGA-II     & $0.00 \pm 0.00$      & $0.00 \pm 0.00$      & -           & $6.88 \pm 0.88$      \\
SMS-EMOA    & $0.00 \pm 0.00$      & $0.00 \pm 0.00$      & -           & $7.62 \pm 0.84$      \\
Pareto-MCTS & $0.00 \pm 0.00$      & $0.00 \pm 0.00$      & -           & $11.62 \pm 1.34$      \\
PLS         & $0.00 \pm 0.00$      & $0.00 \pm 0.00$      & -           & $10.62 \pm 3.23$      \\
MOEA/D      & $0.00 \pm 0.00$      & $0.00 \pm 0.00$      & -           & $12.62 \pm 1.25$      \\
Pareto-NRPA & \bm{$0.39 \pm 0.23$} & \bm{$0.15 \pm 0.10$} & \bm{$7.22 \pm 5.82$} & \bm{$0.12 \pm 0.23$} \\ 
\bottomrule
\end{tabular}
\end{table}

\begin{table}[H]
\centering
\renewcommand{\arraystretch}{1.4}
\caption{Metrics on rc\_204.1}
\label{table:metrics-rc204-1-equal-cpu}
\begin{tabular}{lcccr}
\toprule
Algorithm & Hypervolume & Overall Spread & Spacing & CV \\
\midrule
NSGA-II     & $0.00 \pm 0.00$      & $0.00 \pm 0.00$      & -           & $2.62 \pm 0.98$      \\
SMS-EMOA    & \bm{$0.12 \pm 0.23$} & \bm{$0.12 \pm 0.23$}  & \bm{$3.31 \pm 0.00$} & \bm{$2.25 \pm 0.83$} \\
Pareto-MCTS & $0.00 \pm 0.00$      & $0.00 \pm 0.00$      & -           & $22.12 \pm 0.23$     \\
PLS         & $0.00 \pm 0.00$      & $0.00 \pm 0.00$      & -           & $4.75 \pm 1.46$      \\
MOEA/D      & $0.00 \pm 0.00$      & $0.00 \pm 0.00$      & -           & $4.50 \pm 0.92$      \\
Pareto-NRPA & $0.00 \pm 0.00$      & $0.00 \pm 0.00$      & -           & $10.25 \pm 1.08$ \\ 
\bottomrule
\end{tabular}
\end{table}

Tables \ref{table:metrics-rc204-3-equal-cpu}, \ref{table:metrics-rc201-3-equal-cpu} and \ref{table:metrics-rc204-1-equal-cpu} display metrics for all algorithms with a CPU runtime limited to 240 seconds. Naturally, the number of function evaluations differs greatly under this scenario. For rc\_204.3, which is a very easy and small search space, Pareto Local Search gives the best performances, due to the algorithm's exceptional speed. However, rc\_201.3, which is the hardest instance but with a moderate number of cities, gives back the advantage to Pareto-NRPA. Finally, rc\_204.1, which is the largest instance, is best solved by SMS-EMOA. Remarkably, Pareto-NRPA performs poorly on rc\_204.1 when the search time is constrained: indeed, the large search space increases the complexity of the algorithm, leading to the extremely small number of approximately 1600 function evaluations performed during the time frame. We believe that reducing the computational complexity of Pareto-NRPA could have a very positive effect on performances under a constrained search time.

\end{document}